\documentclass{article}

\usepackage{microtype}
\usepackage{graphicx}
\usepackage{multirow}
\usepackage{subcaption}
\usepackage{booktabs} 
\usepackage{makecell}
\usepackage{pifont}
\usepackage{hyperref}
\usepackage{enumitem}
\usepackage{algorithm}
\usepackage{algorithmic}
\usepackage{amsmath} 
\usepackage{amssymb} 

\usepackage[preprint]{icml2026}

\usepackage{xspace}
\usepackage{amsmath}
\usepackage{amssymb}
\usepackage{mathtools}
\usepackage{amsthm}
\usepackage{soul}
\usepackage[normalem]{ulem}
\usepackage{newtxtext,newtxmath}
\usepackage[table,xcdraw]{xcolor}

\useunder{\uline}{\ul}{}
\usepackage[capitalize,noabbrev]{cleveref}
\newcommand{\name}{DiM\xspace}
\newcommand{\vname}{DiM-V\xspace}
\newcommand{\eg}{\textit{e.g.}\xspace}

\newcommand{\RETURN}{\STATE \textbf{return}}
\theoremstyle{plain}
\newtheorem{theorem}{Theorem}[section]

\theoremstyle{definition}
\newtheorem{definition}[theorem]{Definition}

\theoremstyle{remark}

\usepackage[textsize=tiny]{todonotes}

\icmltitlerunning{Unifying Watermarking via Dimension-Aware Mapping}
\newcommand{\gain}[1]{{\scriptsize\textcolor{green!60!black}{(#1)}}}
\newcommand{\loss}[1]{{\scriptsize\textcolor{red!70!black}{(#1)}}}
\newcommand{\steady}[1]{{\scriptsize\textcolor{gray}{(#1)}}}
\definecolor{lightpurple}{RGB}{250, 248, 255}
\definecolor{myblue}{HTML}{AEC7E8}
\definecolor{rowgray}{gray}{0.95}
\definecolor{iceblue}{HTML}{F0F8FF}

\begin{document}

\twocolumn[
  \icmltitle{Unifying Watermarking via Dimension-Aware Mapping}



  \icmlsetsymbol{equal}{*}

  \begin{icmlauthorlist}
    \icmlauthor{Jiale Meng}{zju}
    \icmlauthor{Runyi Hu}{ntu}
    \icmlauthor{Jie Zhang}{astar}
    \icmlauthor{Zheming Lu}{zju}
    \icmlauthor{Ivor Tsang}{astar}
    \icmlauthor{Tianwei Zhang}{ntu}
  \end{icmlauthorlist}
  
  \icmlaffiliation{zju}{Zhejiang University}
  \icmlaffiliation{ntu}{Nanyang Technological University}
  \icmlaffiliation{astar}{CFAR and IHPC, A$\star$STAR, Singapore}

  \icmlcorrespondingauthor{Jie Zhang}{zhangj6@a-star.edu.sg}

  \icmlkeywords{Machine Learning, ICML}

  \vskip 0.3in
]



\printAffiliationsAndNotice{}  

\begin{abstract}
Deep watermarking methods often share similar encoder–decoder architectures, yet differ substantially in their functional behaviors. We propose DiM, a new multi-dimensional watermarking framework that formulates watermarking as \textbf{a dimension-aware mapping problem}, thereby unifying existing watermarking methods at the functional level.
Under \name, watermark information is modeled as payloads of different dimensionalities, including one-dimensional binary messages, two-dimensional spatial masks, and three-dimensional spatiotemporal structures. We find that the dimensional configuration of embedding and extraction largely determines the resulting watermarking behavior. Same-dimensional mappings preserve payload structure and support fine-grained control, while cross-dimensional mappings enable spatial or spatiotemporal localization.
We instantiate \name in the video domain, where spatiotemporal representations enable a broader set of dimension mappings. Experiments demonstrate that varying only the embedding and extraction dimensions, without architectural changes, leads to different watermarking capabilities, including spatiotemporal tamper localization, local embedding control, and recovery of temporal order under frame disruptions.
\end{abstract}

\section{Introduction}
Deep watermarking is a fundamental technique for ensuring copyright attribution~\cite{zhang2021deep} and content integrity~\cite{neekhara2024facesigns} in visual media. In recent years, the end-to-end training paradigm has provided a unified and flexible framework for deep watermarking~\cite{zhu2018hidden, luo2020distortion}. By adopting encoder–decoder architectures and incorporating noise layers to simulate attack processes, a large body of work has evolved under this paradigm, achieving varying trade-offs between robustness and visual quality~\cite{jia2021mbrs, wu2023sepmark, tancik2020stegastamp}. Despite sharing highly similar network architectures, existing methods are typically designed for specific tasks.
Most prior approaches focus on embedding and extracting one-dimensional binary messages for copyright verification~\cite{Trustmark-ICCV-2025, hu2025robust}. With the growing demand for content authentication and tamper detection, some methods introduce additional fragile watermarks to enable spatial localization~\cite{zhang2024editguard, zhang2025omniguard}, while others formulate localization as a spatially resolved prediction problem at the decoding stage~\cite{hu2025mask, sander2025watermark}.
As a result, different watermarking functionalities are realized through task-specific designs, often without a shared modeling perspective. This observation raises a natural question: \textit{can these seemingly disparate watermarking designs be interpreted within a unified modeling framework? }

\begin{figure}[t]
\centering
\includegraphics[width=0.9\linewidth]{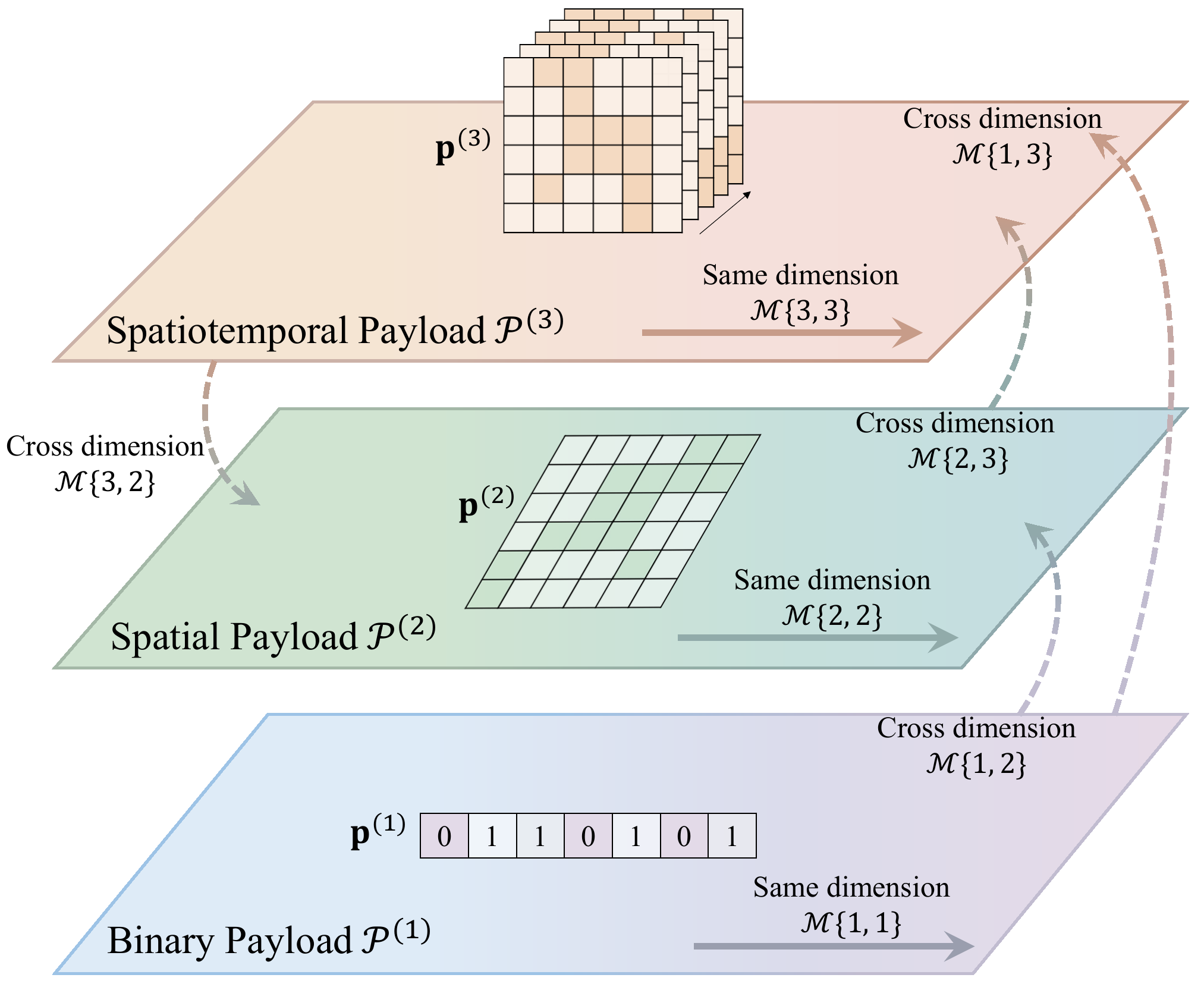}
\caption{
Dimension-aware embedding-extraction mappings across multi-dimensional watermark payloads.
Watermark information is represented at different dimensionalities, and each mapping $\mathcal{M}\{d_e,d_d\}$ is defined by the dimensional relationship between embedding and extraction.
Same-dimensional mappings operate within a single space, while cross-dimensional mappings bridge different dimensional spaces. 
}
\label{fig:flowwork}
\vspace{-1em}
\end{figure}

To answer this question, we propose \textbf{\name}, a unified watermarking framework that formulates watermarking as \textit{a dimension-aware mapping problem.} Within \name, watermark information is modeled as payloads of different dimensionalities, including one-dimensional binary messages, two-dimensional spatial masks, and three-dimensional spatiotemporal structures (see Figure~\ref{fig:flowwork}). Distinct dimensional configurations naturally correspond to different watermarking capabilities. Specifically, when embedding and extraction are performed in the same dimensional space, the watermark enables precise and fine-grained control (\eg, copyright verification). When embedding occurs in a lower-dimensional space and extraction in a higher-dimensional space, the watermark naturally supports coarser-grained functionalities (\eg, tamper localization). Conversely, employing high-dimensional embedding with low-dimensional extraction enables progressive decoupling and recovery of information, effectively reducing decoding complexity and ensuring stable recovery when temporal dependencies in the payload are weak.

Under the unified \name, existing image and video watermarking methods can be systematically understood as specific instances of the proposed framework rather than isolated, task-driven designs. However, prior work on cross-dimensional mappings has been largely restricted to one- and two-dimensional settings. To study more general dimensional configurations, we focus on video watermarking and instantiate \name as a concrete video watermarking approach, denoted as \textbf{\vname}. The inherently high-dimensional spatiotemporal structure of video enables a broader range of embedding and extraction dimension combinations, making it a suitable setting for examining their functional consequences. Guided by the mapping configurations defined in \name, we conduct experiments across multiple dimensional settings. The results demonstrate that, without modifying the network architecture, merely adjusting the embedding and extraction dimensions leads to fundamentally different watermarking capabilities. Beyond conventional copyright protection, \vname fills critical gaps in video watermarking by enabling spatiotemporal tamper localization and fine-grained local embedding. Moreover, by encoding temporal information using high-dimensional payloads, the watermark can explicitly represent frame-order structure and recover the original temporal sequence even when frame order is disrupted. These findings collectively validate the central role of multi-dimensional watermark modeling in the design of the proposed \name framework.

\section{Background}
Our proposed framework is primarily grounded in two domains: image watermarking and video watermarking, which we briefly review from the perspective of \textit{how watermark information is represented and extracted}.

\noindent\textbf{Image Watermarking.}
Existing image watermarking methods~\cite{petrov2025we, hu2024supermark,bui2023rosteals,wen2023tree} can be broadly categorized by how watermark information is embedded and extracted. The most common paradigm performs global embedding and extraction of a one-dimensional binary message, primarily targeting robust copyright verification under common distortions, as exemplified by Robust-Wide~\cite{hu2025robust}, StegaStamp~\cite{tancik2020stegastamp}, and MBRS~\cite{jia2021mbrs}. To support tamper localization, some approaches retain global watermarking but embed additional fragile signals whose degradation reveals manipulated regions, such as EditGuard~\cite{zhang2024editguard} and OmniGuard~\cite{zhang2025omniguard}. Another line of work achieves localization by reformulating extraction as a spatially resolved prediction problem.  For example, WAM~\cite{sander2025watermark} embeds a global one-dimensional message while performing pixel-wise extraction to identify tampered regions; MaskWM~\cite{hu2025mask} further extends this paradigm with a mask-guided design, supporting global embedding with local extraction. Its MaskWM-ED variant additionally incorporates two-dimensional masks at the embedding stage to enable local embedding. 
These methods demonstrate that \textit{different watermarking tasks require different information representations. However, even for the same task, existing approaches adopt substantially different watermark representations, which are largely introduced in an ad hoc manner without a unified formulation.}

\noindent\textbf{Video Watermarking.}
Compared to image watermarking, video watermarking~\cite{luo2023dvmark, zhang2024v2a,jiang2025videomarkbench,souvcek2025pixel,ji2026dinvmark,ye2023itov} methods remain focused on global embedding.
RivaGAN~\cite{zhang2019robust} employs attention mechanisms and adversarial training for robust video watermarking; REVMark~\cite{zhang2023novel} improves robustness to compression via temporal alignment and H.264-aware perturbations; VideoSeal~\cite{fernandez2024videoseal} extends image watermarking methods to videos through temporal watermark propagation.
Despite these advances, current video watermarking methods suffer from two key limitations. \emph{First,} watermark information is almost exclusively modeled as a global one-dimensional message, with little explicit modeling of spatial or temporal localization, thus restricting their capability for tasks such as tamper localization. \emph{Second,} temporal order is insufficiently modeled, which constitutes a critical limitation for video watermarking applications~\cite{hu2025videoshield}. When frames are reordered, inserted, or shuffled, existing approaches struggle to encode or recover temporal structure.

Motivated by these observations, we propose a unified multi-dimensional perspective for watermark information modeling. This perspective organizes existing methods by dimensionality of the watermark representation and naturally exposes the limitations of current video watermarking designs, while enabling explicit modeling of spatial locality and temporal structure.

\section{\name: Unified Watermarking Framework}
In this section, we formulate deep watermarking as a dimension-aware mapping problem.
The central premise of \name is that watermarking functionality is governed
by the dimensionality with which watermark information is represented during embedding and recovered during extraction.
Unlike conventional approaches that impose a predefined watermark form (\eg, fixed-length one-dimensional bit sequences), we model watermark information as a random variable whose functional capacity is determined by the dimensionality of its representation.

\subsection{Multi-dimensional Payload Space}
\label{sec:payload_space}
We first define a set of multi-dimensional payload spaces, which serve as the representational foundation of \name.
Let $\mathcal{P}^{(d)}$ denote a watermark payload defined in a $d$-dimensional representation space.
We consider three payload spaces corresponding to $d \in \{1,2,3\}$.

\begin{definition}[1D Binary Payload]
    The 1D payload, denoted as $\mathbf{p}^{(1)} \in \mathcal{P}^{(1)}$, represents global, permutation-invariant binary information that encodes binary identity or ownership. We define the 1D payload space as:
    \begin{equation}
        \mathcal{P}^{(1)} = \{0, 1\}^L 
    \end{equation}
    where $L$ is the length of the binary payload.
\end{definition}

\begin{definition}[2D Spatial Payload] 
The 2D payload $\mathbf{p}^{(2)} \in \mathcal{P}^{(2)}$ represents spatially localized structural information that captures region-level or geometric constraints.
\begin{equation}
\mathcal{P}^{(2)} = \mathbb{R}^{H \times W \times C_p}
\end{equation}
where $H, W$ are spatial dimensions and $C_p$ denotes the number of payload channels.
\end{definition}

\begin{definition}[3D Spatiotemporal Payload] 
The 3D payload $\mathbf{p}^{(3)} \in \mathcal{P}^{(3)}$ represents spatiotemporal structural information defined over space-time volumes.
\begin{equation}
\mathcal{P}^{(3)} = \mathbb{R}^{T \times H \times W \times C_p}
\end{equation}
where $T$ is the temporal dimension. 
\end{definition}
These payloads do not correspond to distinct tasks. Instead, they represent unified information primitive instantiated across a spectrum of representational dimensionalities.
Figure~\ref{fig:flowwork} illustrates the relationships between payload spaces and the corresponding same- and cross-dimensional mappings.

\subsection{Dimension-aware Mapping}
\label{sec:dimension_mapping}
Let $x \in \mathcal{X}$ denote the host signal, such as an image or a video.
We define the watermark embedding and extraction processes as dimension-aware functions:
\begin{align}
\mathcal{E}_{\theta} &: \mathcal{X}_{\text{orig}} \times \mathcal{P}^{(d_e)} \rightarrow \mathcal{X}_{\text{wm}}, \\
\mathcal{D}_{\phi} &: \mathcal{X}_{\text{wm}} \rightarrow \hat{\mathcal{P}}^{(d_d)},
\end{align}
where $\mathcal{E}_{\theta}$ represents the embedding network (Encoder) with trainable parameters $\theta$, taking the host $x$ and a payload of dimension $d_e$ as inputs. Similarly, $\mathcal{D}_{\phi}$ represents the extraction network (Decoder) with parameters $\phi$, designed to recover a payload of dimension $d_d$.
Let $\mathcal{A}$ denotes an unknown distortion or attack process, the entire end-to-end pipeline can then be expressed as a unified mapping:
\begin{equation}
\mathcal{M}\{d_e, d_d\}: \mathcal{P}^{(d_e)}
\;\xrightarrow{\;\mathcal{E}_{\theta}\;}\;
\mathcal{X}_{\text{wm}}
\;\xrightarrow{\;\mathcal{A}\;}\;
\tilde{\mathcal{X}}_{\text{wm}}
\;\xrightarrow{\;\mathcal{D}_{\phi}\;}\;
\hat{\mathcal{P}}^{(d_d)},
\end{equation}

A key property of \name is the relationship between the embedding dimension $d_e$ and the extraction dimension $d_d$.

\noindent\textbf{\boldmath$\mathcal{M}\{d_e=d_d\}$: Same-dimensional Mapping.} 
When embedding and extraction are performed in the same dimension, the extraction process is simplified to the estimation of the original payload:
\begin{equation}
\hat{\mathcal{P}}^{(d_e)} \approx \mathcal{P}^{(d_e)}.
\end{equation}
This regime preserves the payload structure, thereby allowing for fine-grained control (\eg, copyright verification).

\begin{table}[t]
\caption{Existing image and video watermarking methods under the DiMap framework. Methods are grouped by embedding-extraction regimes $\mathcal{M}\{d_e,d_d\}$. Most prior work focuses on same-dimensional mappings, with limited coverage of cross-dimensional regimes.}
\vspace{-1em}
\label{tab:mappings-in-prior-work}
\begin{center}
\begin{small}
\begin{sc}
\setlength{\tabcolsep}{1pt}
\begin{tabular}{cccc}
\toprule
\boldmath$ \mathcal{M}\{d_e,d_d\} $ & \boldmath$d_e=1$ & \boldmath$d_e=2$ & \boldmath$d_e=3$ \\
\midrule
\rowcolor{rowgray}
\boldmath$d_d=1$ &\cellcolor{rowgray}
\makecell{Hidden, MaskWM, WAM\\ TrustMark, VideoSeal,\\ REVMark, RivaGAN, \\EditGuard, OmniGuard, \\Robust-Wide, \textit{etc}. } &
N/A &
N/A \\
\boldmath$d_d=2$ &
MaskWM-D, WAM &
\makecell[c]{MaskWM-ED,\\EditGuard, \\OmniGuard} &
N/A \\
\rowcolor{rowgray}
\boldmath$d_d=3$ &
N/A &
N/A &
N/A \\
\bottomrule
\end{tabular}
\end{sc}
\end{small}
\end{center}
\vspace{-0.5em}
\end{table}

\noindent\textbf{\boldmath$\mathcal{M}\{d_e\!\neq\!d_d\}$: Cross-dimensional Mapping.} 
When $d_e \neq d_d$, watermark extraction induces a \emph{cross-dimensional projection}. We further distinguish two regimes.
\begin{itemize}[nosep]
    \item {\textbf{Low-to-High Mapping} \boldmath$\mathcal{M}\{d_e<d_d\}$}. Low-dimensional payloads are embedded and recovered in a higher-dimensional representation space. Extraction can be interpreted as
    \begin{equation}
    \hat{\mathcal{P}}^{(d_d)} \in \Pi^{\uparrow}_{d_e \rightarrow d_d} \left( \mathcal{P}^{(d_e)} \right),
    \end{equation}
    where $\Pi^{\uparrow}_{d_e \rightarrow d_d}$ denotes a structure-expanding projection. This regime gives rise to spatial or spatiotemporal localization binary message. 

    \item {\textbf{High-to-Low Mapping} \boldmath$\mathcal{M}\{d_e\!>\!d_d\}$}.
    High-dimensional payloads are embedded, while extraction targets a lower-dimensional representation:
    \begin{equation}
    \hat{\mathcal{P}}^{(d_d)} 
    = \Pi^{\downarrow}_{d_e \rightarrow d_d}
    \left( \mathcal{P}^{(d_e)} \right), 
    \end{equation}
    where $\Pi^{\downarrow}_{d_e \rightarrow d_d}$ denotes a structure-decoupling projection. Our empirical analysis indicates that this mapping becomes necessary when cross-dimensional coherence is weak, under which direct high-dimensional extraction is ill-conditioned.
\end{itemize}

From a functional perspective, \name provides a unified view of existing image and video watermarking methods.
Table~\ref{tab:mappings-in-prior-work} summarizes prior approaches under this framework. It can be observed that most existing methods focus on the mapping $\mathcal{M}\{1,1\}$.
In the following section, we instantiate \name in the video domain and examine multiple embedding-extraction regimes beyond those covered by prior work.

\section{\vname: \name for Video Watermarking}
\label{sec:video method}

In this section, we instantiate \name in the context of video watermarking, and refer to this concrete instantiation as \vname.
Our goal is not to introduce a task-specific video watermarking architecture, but to demonstrate how different dimension-aware mappings naturally give rise to diverse watermarking functionalities. We first ground the abstract payload spaces defined in Section~\ref{sec:payload_space} in concrete video representations, and then show how varying only the dimensional configuration of embedding and extraction induces distinct behaviors without architectural modifications.

\subsection{Payload Instantiation}
\label{sec:payload_instantiation}

We consider a video clip $\mathbf{V}_{\text{orig}} \in \mathbb{R}^{T \times H \times W \times 3}$, where $T$ denotes the temporal length and $H,W$ denote the spatial resolution.
Following the multi-dimensional payload formulation in Section~\ref{sec:payload_space}, we instantiate three payload representations for video watermarking.

\noindent\textbf{1D Binary Payload $\mathbf{W}$.}
The binary payload is instantiated as a randomly sampled binary message of length $L$, denoted as $\mathbf{W} \in \mathcal{P}^{(1)} = \{0,1\}^{L}$.
It encodes global, permutation-invariant information, such as ownership or identity, and is independent of the spatial or temporal structure of the video.

\noindent\textbf{2D Spatial Payload $\mathbf{M}^{(2)}$.}
The spatial payload is instantiated as a spatial mask
$\mathbf{M}^{(2)} \in \mathcal{P}^{(2)}=\mathbb{R}^{H \times W \times C_p^{\mathbf{M}^{(2)}}}$, which represents region-level structural information.
Following MaskWM~\cite{hu2025mask} and WAM~\cite{sander2025watermark}, we adopt the mask generation strategy of LaMa~\cite{suvorov2022resolution}, constructing four types of masks: full masks, rectangular masks, irregular masks, and segmented masks. As segmented masks are defined on a per-frame basis and inherently indexed in time, we randomly sample a single frame-level mask and treat it as a static 2D spatial payload.

\noindent\textbf{3D Spatiotemporal Payload $\mathbf{M}^{(3)}$.}
The 3D payload is instantiated as a spatiotemporal mask tensor
$\mathbf{M}^{(3)} \in \mathcal{P}^{(3)}=T \times H \times W \times C_p^{\mathbf{M}^{(3)}}$ which encodes temporally varying spatial structures across video frames. Consistent with the 2D setting, we consider the same four categories of masks. For rectangular and irregular masks, we employ a simple yet effective temporal mask-shifting strategy that propagates a single spatial mask across frames,  thereby emulating object-tracking–like behavior along the temporal dimension (see Appendix~\ref{sec: mask shift}).
In the specific case of full masks, setting $C_p^{\mathbf{M}^{(3)}} = 1$ results in identical masks across all frames and provides no temporal differentiation. More generally, conventional mask designs lack the capacity to encode or infer temporal order because they do not utilize an explicit, permutation-invariant representation of frame identity. 

\begin{figure}
\centering
\includegraphics[width=1.0\linewidth]{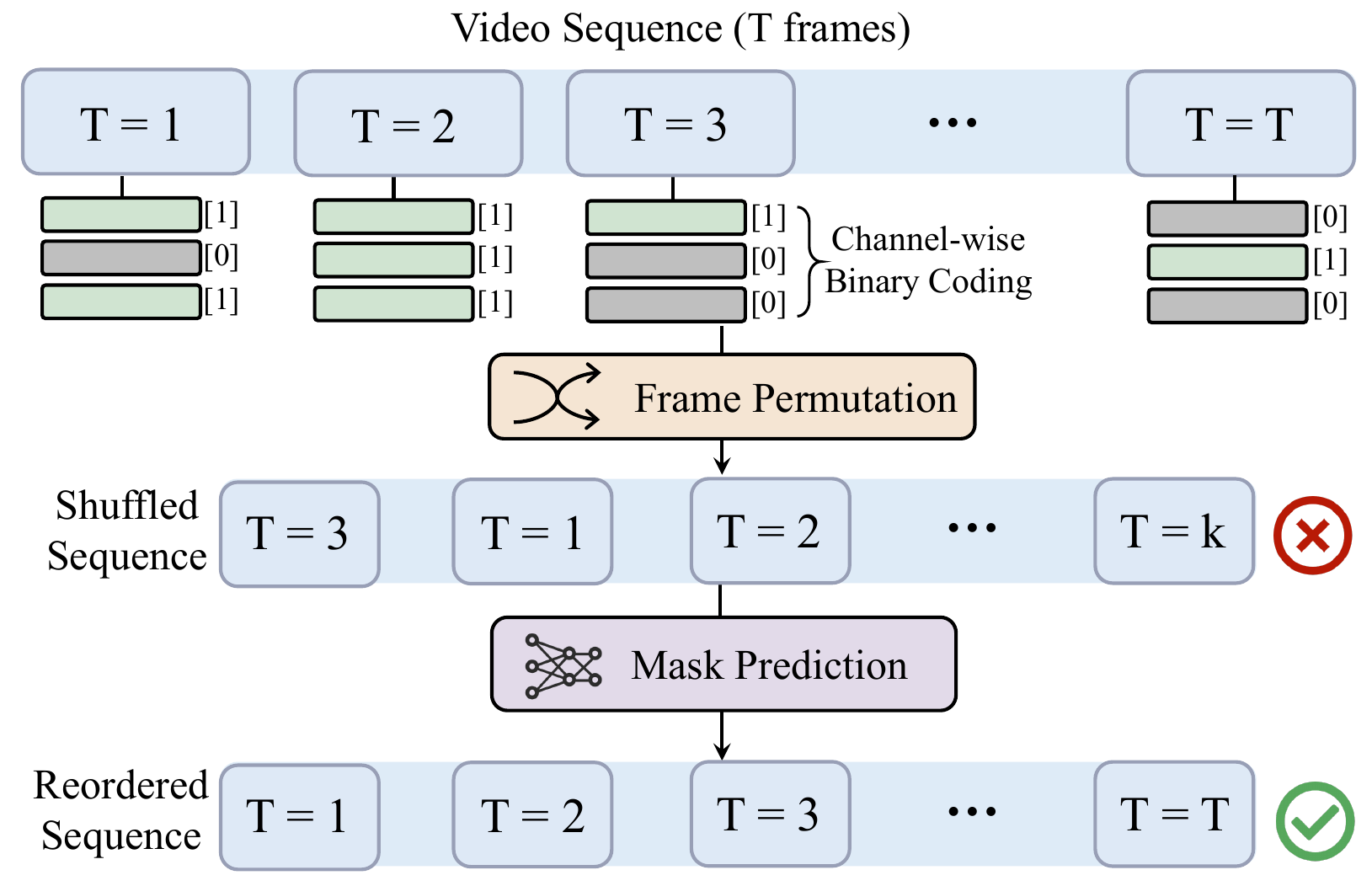}
\caption{Illustration of the proposed multi-channel mask encoding for spatiotemporal payloads. Each frame is assigned a distinct, permutation-invariant binary code along the channel dimension, where each channel map is spatially uniform within a frame (all-0 or all-1). This encoding scheme enables frame-wise localization and temporal order recovery under arbitrary frame permutations.}
\label{fig:multibitmask}
\vspace{-1.5em}
\end{figure}

To address this limitation, we introduce a \textit{multi-channel mask encoding mechanism} where $C_p^{\mathbf{M}^{(3)}} > 1$.
In this scheme, each frame is assigned a distinct, permutation-invariant binary code along the channel dimension to explicitly encode its identity.
We strictly exclude all-zero codes to prevent ambiguity with the unmasked state.
As illustrated in Figure~\ref{fig:multibitmask}, this representation enables precise frame-wise localization and supports reliable temporal order recovery even under arbitrary frame permutations.

\subsection{Input Construction}
\label{sec:input_tensor}
All dimension-aware mappings in \name share a unified input representation.
Specifically, the input tensor $\mathbf{T}_{\text{in}} \in \mathbb{R}^{C_{\text{in}} \times T \times H \times W}$ is constructed by channel-wise concatenation of watermark-related features and the host video $\mathbf{V}_{\text{orig}} \in \mathbb{R}^{3 \times T \times H \times W}$.
The watermark-related features consist of two parts: a feature representation derived from the 1D message $\mathbf{W}$, and an optional spatial or spatiotemporal mask payload, depending on the mapping.
The 1D message $\mathbf{W}$ is always included, as it serves as the core carrier of watermark identity and ensures copyright verification across all mappings.
When applicable, a 2D spatial mask $\mathbf{M}^{(2)}$ or a 3D spatiotemporal mask $\mathbf{M}^{(3)}$ is additionally incorporated to induce locality or temporal structure.
Below, we describe how $\mathbf{T}_{\text{in}}$ is instantiated under representative mappings and analyze the functional behaviors induced by each dimensional configuration.

\noindent\textbf{Same Dimension \boldmath$\mathcal{M}\{3,3\}$.}
1D message $\mathbf{W}$ is first transformed by a message translator into a spatiotemporal feature tensor
$\mathbf{T}_{\text{msg}} \in \mathbb{R}^{C_\text{tp} \times T \times H \times W}$, which is then concatenated with a 3D mask $\mathbf{M}^{(3)}$ and the original video $\mathbf{V}_{\text{orig}}$ along the channel dimension, yielding
$\mathbf{T}_{\text{in}} =
\mathrm{Concat}\big(
\mathbf{V}_{\text{orig}},\;
\mathbf{T}_{\text{msg}},\;
\mathbf{M}^{(3)}
\big)$.
This paradigm supports both global and local embedding of the message. Global embedding refers to embedding the payload across the entire video, while local embedding corresponds to embedding the payload within a specific object or region of interest, such as along an object trajectory.

\noindent\textbf{Cross Dimension \boldmath$\mathcal{M}\{1,3\}$.}
In this setting, the input tensor is constructed as
$\mathbf{T}_{\text{in}} =
\mathrm{Concat}\big(
\mathbf{V}_{\text{orig}},\;
\mathbf{T}_{\text{msg}}
\big)$.
Embedding a low-dimensional payload while extracting a spatiotemporal representation induces a structure-expanding projection, enabling effective localization of tampering events such as frame removal or object deletion.

\noindent\textbf{Cross Dimension \boldmath$\mathcal{M}\{2,3\}$.} A 2D payload $\mathbf{M}^{(2)}$ is replicated along the temporal dimension to form a mask sequence $\mathbf{M}^{(3)}$ aligned with the video length. The resulting input tensor is given by 
$\mathbf{T}_{\text{in}} =
\mathrm{Concat}\big(
\mathbf{V}_{\text{orig}},\;
\mathbf{T}_{\text{msg}},\;
\mathbf{M}^{(3)}
\big)$.
Compared to $\mathcal{M}\{1,3\}$, this mapping introduces explicit spatial constraints, enabling controlled local embedding while retaining spatiotemporal localization capability.

\noindent\textbf{Cross Dimension \boldmath$\mathcal{M}\{3,2\}$.}
The input tensor is the same as $\mathcal{M}\{3,3\}$. When high-dimensional spatiotemporal payloads exhibit weak cross-frame coherence, direct high-dimensional extraction becomes ill-conditioned.
Empirically, increasing the channel $C_p^{\mathbf{M}^{(3)}}$ exacerbates this effect by weakening temporal correlations across frames.
To address this challenge, we adopt a decoupled modeling strategy in which spatiotemporal masks are predicted independently on a per-frame basis.
This decomposition substantially reduces prediction complexity and improves both stability and accuracy of mask recovery and localization. 

Regarding the outputs, $\mathcal{M}\{3,3\}$, $\mathcal{M}\{1,3\}$, and $\mathcal{M}\{2,3\}$ produce a 1D message together with a 3D mask aligned with the video volume.
In contrast, $\mathcal{M}\{3,2\}$ yields a 1D message and a sequence of frame-wise 2D masks aligned with the temporal dimension of the input video.

\subsection{Unified Watermarking Pipeline}
Across all mappings, we employ a unified network architecture and training pipeline equipped with a shared noise layer.
Specifically, we adapt the MaskWM framework to the spatiotemporal domain and structure the pipeline into four stages: watermark embedding, watermark masking, watermark extraction, and mask prediction.
Architectural and training details are provided in Appendix~\ref{sec:training_pipeline}.

\begin{table*}[!t]
\caption{Comparison of imperceptibility and robustness against baseline methods in terms of global watermarking in video scenarios. The symbols \ding{109} and \ding{111} represent image and video watermarking methods, respectively. Bits denotes the number of embedded bits. Qualitative capabilities are indicated by checkmarks: L-E, L-X, and Loc denote local embedding, local extraction, and tamper localization, respectively. The best and second-best results are highlighted in \textbf{bold} and {\ul underlined}, respectively.}
\vspace{-0.5em}
\label{tab:global-same}
\centering
\begin{center}
\begin{small}
\begin{sc}
\setlength{\tabcolsep}{2pt}
\begin{tabular}{lccccccccccc}
\toprule
\multirow{2}{*}{\textbf{Method}} & \multirow{2}{*}{\textbf{Bits}} & \multirow{2}{*}{\textbf{PSNR $\uparrow$} } & \multirow{2}{*}{\textbf{SSIM $\uparrow$}} & \multirow{2}{*}{\textbf{Clean $\uparrow$}}&\multicolumn{2}{c}{\textbf{Common Distortions $\uparrow$}}& \multicolumn{2}{c}{\textbf{Video Distortions $\uparrow$}}& \multicolumn{3}{c}{\textbf{Capabilities}}\\
\cmidrule(lr){7-8} \cmidrule(lr){9-10} 
 &  &  &  &  & Valuemetric & Geometric & Frame-level &  Compression & L-E & L-X & Loc \\
 \midrule
\multicolumn{12}{c}{\cellcolor[HTML]{EFEFEF}\textit{\textbf{Same-dimensional Mapping} \boldmath$\mathcal{M}\{d_e=d_d\}$}}\\
\ding{109} MaskWM-ED & 64 & 39.35 & 0.9704 & \textbf{100} & \textbf{100} & 99.92 & 96.88 & 54.47 &\ding{51}&\ding{51}&\ding{51}\\
\ding{109} OmniGuard & 100 & 39.17 & 0.9759 & {\ul 99.98} & 92.68 & 71.13 & 96.89 & 91.71 &\ding{55}&\ding{55}&\ding{51}\\
\ding{109} TrustMark & 100 & 40.92 & 0.9865 & 99.96 & 95.94 & 78.47 & 97.09 & 89.88 &\ding{55}&\ding{55}&\ding{55}\\
\ding{109} Robust-Wide & 64 & 41.82 & 0.9904 & \textbf{100} & 99.14 & 50.42 & 96.88 & \textbf{99.91} &\ding{55}&\ding{55}&\ding{55}\\
\ding{111} VideoSeal & 96 & \textbf{53.86} & \textbf{0.9988} & 98.54 & 81.39 & 97.28 & {\ul 98.68} & 54.12 &\ding{55}&\ding{55}&\ding{55}\\
\ding{111} RivaGAN & 32 & 40.54 & 0.9793 & 99.82 & 95.95 & 83.06 & 97.29 & 70.27 &\ding{55}&\ding{55}&\ding{55}\\
\ding{111} REVMark & 96 & 37.79 & 0.9905 & 99.97 & 96.69 & 57.25 & 96.63 & 88.22 &\ding{55}&\ding{55}&\ding{55}\\
\rowcolor{iceblue} 
 & 64 & 43.17 & 0.9902 & \textbf{100} & \textbf{100} & \textbf{100} & \textbf{100} & {\ul 98.58} &\ding{51}&\ding{51}&\ding{51}\\
\rowcolor{iceblue}
 & 96 & 42.36 & 0.9855 & \textbf{100} & \textbf{100} & 99.73 & \textbf{100} & 90.11&\ding{51}&\ding{51}&\ding{51}\\
\rowcolor{iceblue} 
\multirow{-3}{*}{\ding{111} \textbf{\vname $\mathcal{M}\{3,3\}$}} & 128 & 41.61 & 0.9847 & \textbf{100} & \textbf{100} & 98.91 & \textbf{100} & 92.99 &\ding{51}&\ding{51}&\ding{51}\\
\midrule
\multicolumn{12}{c}{\cellcolor[HTML]{EFEFEF}\textit{\textbf{Cross-dimensional Mapping} \boldmath$\mathcal{M}\{d_e<d_d\}$}}\\
\midrule
\ding{109} WAM & 32 & 37.18 & 0.9643 & \textbf{100} & {\ul 99.90} & 99.43 & 96.88 & 71.42 &\ding{55}&\ding{51}&\ding{51}\\
\ding{109} MaskWM-D & 64 &38.84&0.9696&\textbf{100}&\textbf{100}&{\ul 99.96}& 96.88&60.95&\ding{55}&\ding{51}&\ding{51}\\
\rowcolor{iceblue}
\ding{111} \textbf{\vname} $\mathcal{M}\{1,3\}$& 64 & {\ul 43.88} & {\ul 0.9919} & \textbf{100} & \textbf{100} & \textbf{100} & \textbf{100} & 98.04&\ding{55}&\ding{51}&\ding{51}\\
\rowcolor{iceblue}
\ding{111} \textbf{\vname} $\mathcal{M}\{2,3\}$& 64  & 43.60 & 0.9905 & \textbf{100} & \textbf{100} & \textbf{100} & \textbf{100} & 92.01&\ding{51}&\ding{51}&\ding{51}\\
\bottomrule
\end{tabular}
\end{sc}
\end{small}
\end{center}
\end{table*}

\begin{figure*}[!h]
\centering
\includegraphics[width=1.00\linewidth]{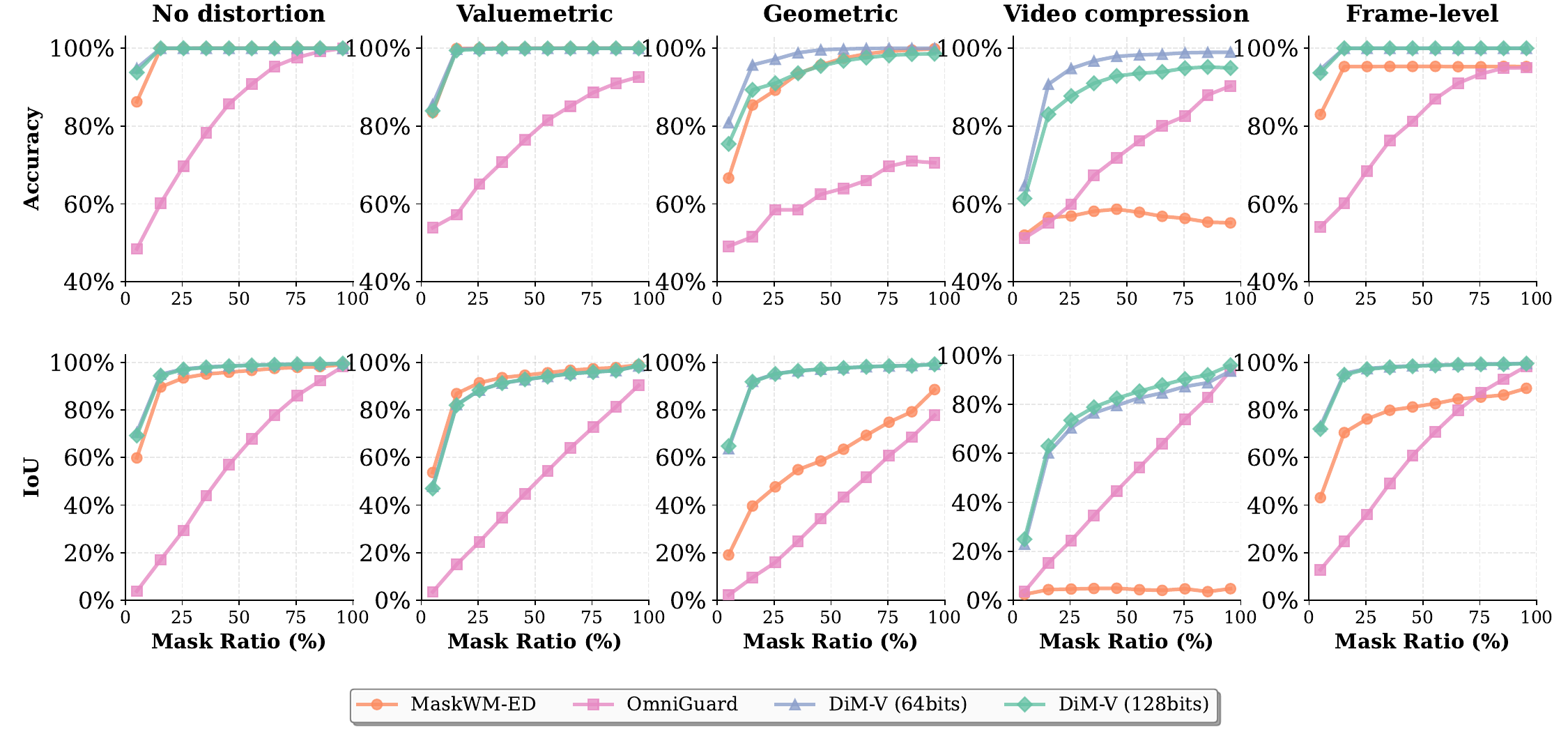}
\caption{
Comparison with baseline methods on local extraction (top) and tamper localization (bottom) in \emph{the same-dimension mapping} setting.
Results are evaluated under five distortion scenarios, and we select the average value for each interval’s ratios to stand for the interval (\eg, 5\% represents the average over the 1--10\% interval).}
\label{fig:local-same}
\end{figure*}

\section{Experiments}
\subsection{Experimental Setup}
\noindent\textbf{Datasets.} For all mapping configurations $\mathcal{M}\{d_e,d_d\}$, we train our models on more than 12{,}000 videos from the SA-V dataset~\cite{ravisam}. 
Training details are provided in Appendix~\ref{sec: training details}.
Evaluation is conducted on two subsets.
The \emph{global} subset contains 1{,}000 randomly sampled videos.
The \emph{local} subset is stratified into ten bins according to the ratio of masked area to the full video volume (0--10\%, \dots, 90--100\%), with 500 videos randomly sampled per bin.
All videos have a resolution of $256 \times 256$ with $T=8$ frames.

\noindent\textbf{Metrics.} 
Visual quality is assessed by PSNR and SSIM.
Robustness is measured by Bit Accuracy under four categories of distortions: valuemetric, geometric, frame-level, and video compression.
Valuemetric distortions include Gaussian noise, Gaussian blur, salt-and-pepper noise, and median filtering.
Geometric distortions include rotation, perspective transformation, and horizontal flipping.
Frame-level distortions include frame replacement, frame dropping, frame insertion, and temporal shuffling.
Video compression is evaluated using H.264.
Results are averaged across distortions within each category.
Localization performance is measured by Intersection-over-Union (IoU) between predicted and ground-truth (gt) watermark regions. For the multi-channel setting, we report the mean IoU (mIoU) by averaging masks with distinct binary codes.
Detailed distortion settings are provided in Appendix~\ref{sec: test distortion}. Efficiency are reported in Appendix~\ref{sec:efficiency}.

\noindent\textbf{Baselines.} We compare with nine recent open-source watermarking methods.
Video watermarking baselines include VideoSeal~\cite{fernandez2024videoseal}, Rivagan~\cite{zhang2019robust}, and RevMark~\cite{zhang2023novel}. 
Image watermarking baselines include MaskWM-D, MaskWM-ED~\cite{hu2025mask}, WAM~\cite{sander2025watermark}, OmniGuard~\cite{zhang2025omniguard}, TrustMark~\cite{Trustmark-ICCV-2025}, and Robust-Wide~\cite{hu2025robust}. 
Image-based methods are evaluated using a frame-wise protocol inspired by VideoSeal, embedding the watermark into each frame and performing frame-level extraction and evaluation, with results averaged to obtain video-level performance. For methods that do not support the target resolution, we adopt the scaling strategy of TrustMark to match our setup.

\subsection{Same-dimensional Mapping: \boldmath$\mathcal{M}\{d_e = d_d\}$}
\noindent\textbf{Global 1D Embedding and Extraction.}
Table~\ref{tab:global-same} presents a comparative evaluation of imperceptibility, robustness, and functional capabilities under the same-dimensional mappings. Regardless of the embedding capacity, \vname demonstrates superior robustness against valuemetric, geometric, and frame-level distortions. Although Robust-Wide shows a slight advantage under compression due to its editing-oriented optimization, \vname achieves a favorable balance between robustness and imperceptibility. We also discuss strategies for improving compression robustness in Appendix~\ref{sec: compression improve}, where notable gains are observed. Beyond quantitative performance, \vname uniquely supports a combination of local embedding, local extraction, and tamper localization within a single configuration. Such functionality is largely absent from prior methods operating under the same-dimensional mapping regime.

\begin{figure*}
    \centering
    \includegraphics[width=1.00\linewidth]{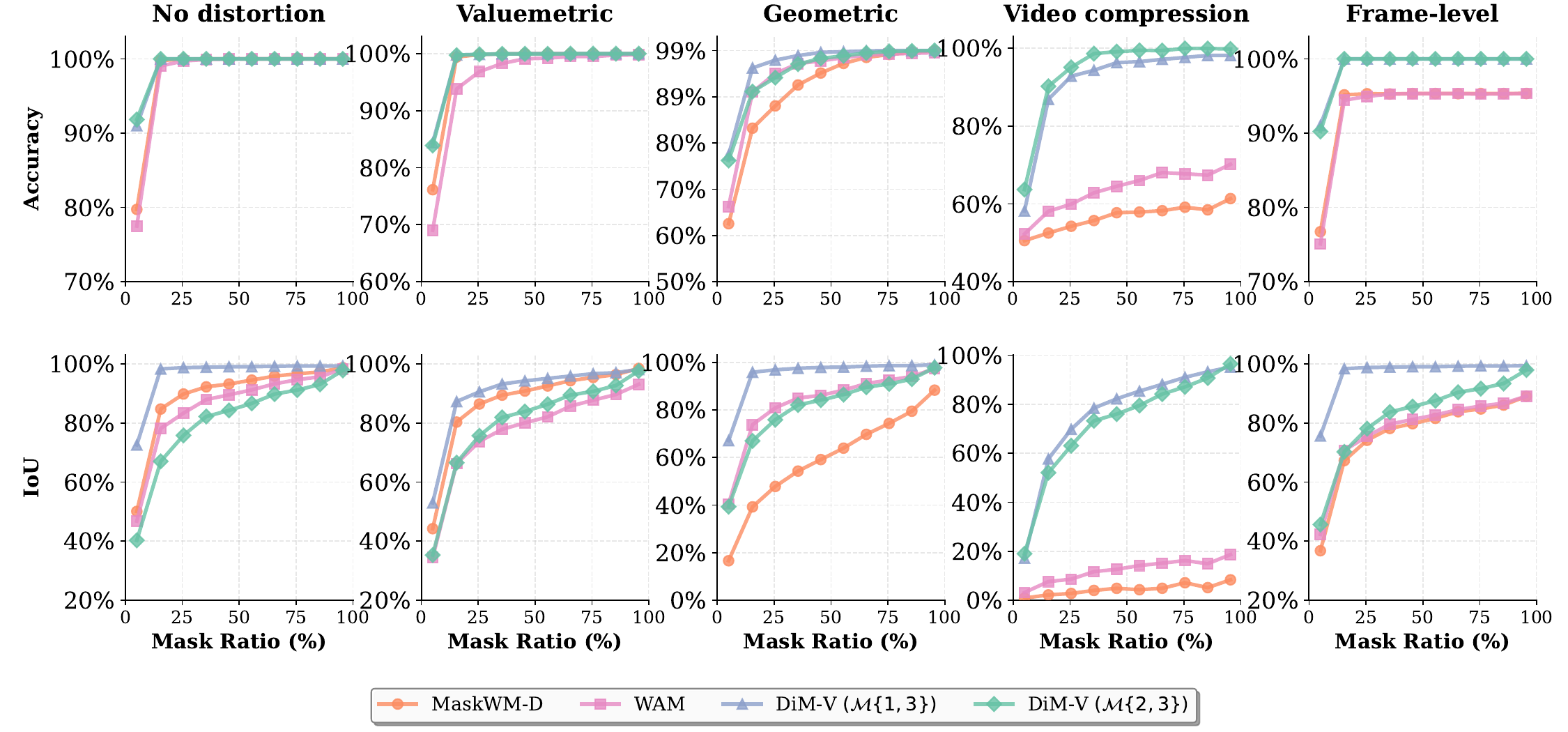}
    \caption{Comparison with baseline methods on local extraction (top) and tamper localization (bottom) in the \emph{cross-dimension mapping} setting. Results are evaluated under five distortion scenarios, and we select the average value for each interval’s ratios to stand for the interval (\eg, 5\% represents the average over the 1--10\% interval).}
    \label{fig:local-cross}
\end{figure*}

\noindent\textbf{Localized Extraction and Localization Performance.}
To assess these capabilities in the same-dimensional mapping, we compare against MaskWM-ED and OmniGuard, which support the $\mathcal{M}\{2,2\}$ regime. As shown in Figure~\ref{fig:local-same}, \vname consistently achieves higher local extraction accuracy across all mask ratios and payload sizes.
The performance gap becomes more pronounced under geometric and video-specific distortions, where baseline methods exhibit unstable extraction.
With respect to localization performance, \vname remains robust across most distortions, with only a modest IoU decrease under valuemetric noise, representing trade-off given the overall robustness and functionality achieved.

\subsection{Low-to-High Mapping: \boldmath$\mathcal{M}\{d_e<d_d\}$}
\noindent\textbf{Global 1D Embedding and Extraction.}
As shown in Table~\ref{tab:global-same}, under the cross-dimensional mapping setting, our method significantly outperforms baselines such as WAM, improving compression robustness by over 20\% while ensuring accurate extraction across other distortion types. Moreover, we demonstrate significantly higher imperceptibility than WAM and MarkWM, realizing an optimal trade-off between robustness and visual fidelity.

\begin{figure}[!t]
    \centering
    \includegraphics[width=\linewidth]{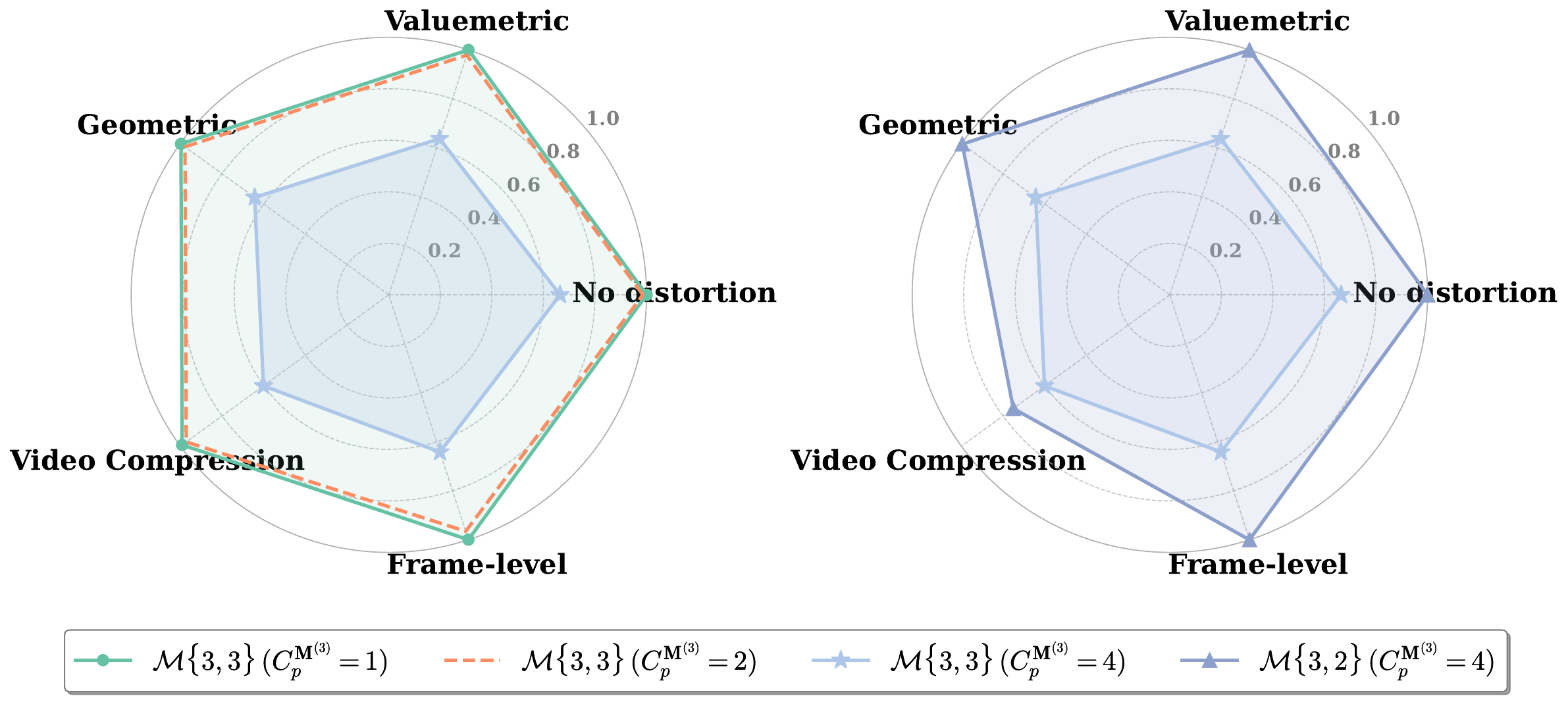}
    \caption{Effect of mask channel count and dimensional mapping on global mIoU under diverse distortions.
Comparing $\mathcal{M}\{3,3\}$ and $\mathcal{M}\{3,2\}$ reveals that increasing mask dimensionality degrades volumetric extraction, while high-to-low mapping restores robustness under multi-channel payloads.}
    \label{fig:channel_compare}
\end{figure}

\begin{figure*}[!t]
    \centering
    \includegraphics[width=1.00\linewidth]{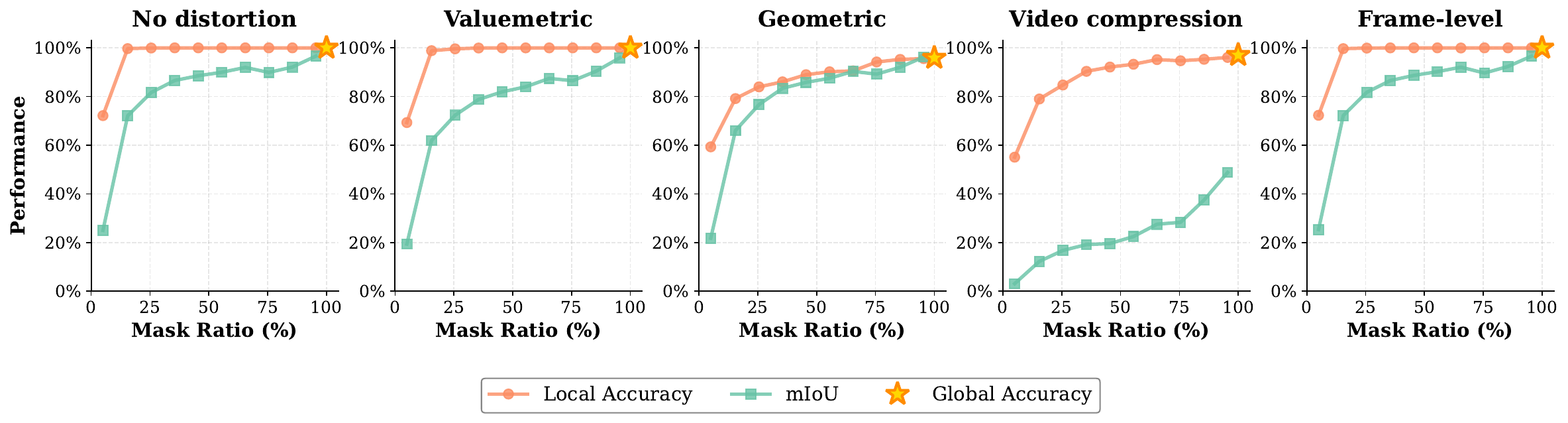}
    \caption{Evaluation of global and local watermark extraction accuracy alongside tamper localization performance under the \emph{multi-channel mask design}. Results are reported under five distortion scenarios, and we select the average value for each interval’s ratios to stand for the interval (\eg, 5\% represents the average over the 1--10\% interval).}
    \label{fig:mc4_acc_miou}
\end{figure*}

\begin{figure*}
    \centering
    \includegraphics[width=1.00\linewidth]{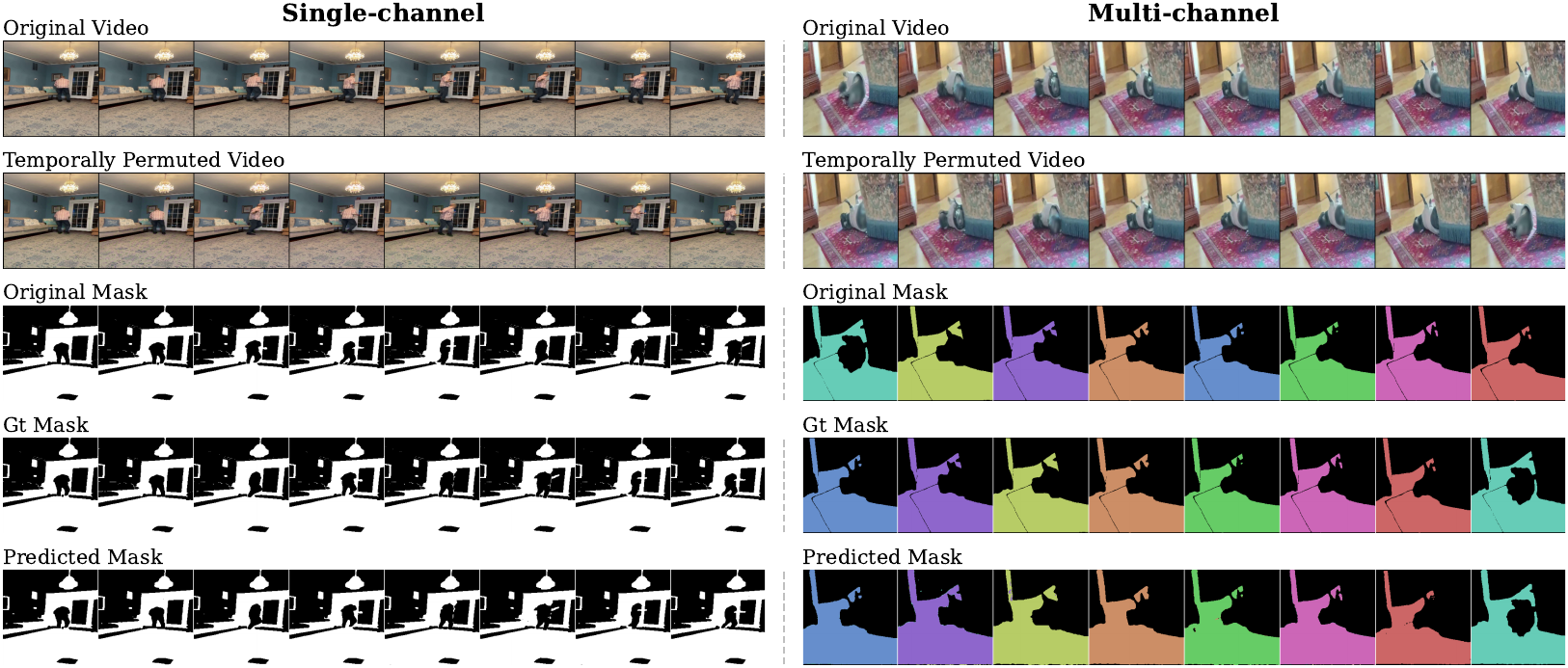}
    \caption{Visualization of single-channel and multi-channel mask representations under temporal permutation.
Both methods enable spatial tamper localization. The multi-channel design assigns distinct encoded masks to different frames, visualized using different colors, allowing frame-specific identity inference and implicitly supporting temporal order localization.}
    \label{fig:merge_multi}
\end{figure*}
\noindent\textbf{Localized Extraction and Localization Performance.}
To evaluate structure-expanding mappings, we compare \vname under $\mathcal{M}\{1,3\}$ and $\mathcal{M}\{2,3\}$ with MaskWM-D and WAM.
As shown in Figure~\ref{fig:local-cross}, $\mathcal{M}\{1,3\}$ achieves strong localization and extraction robustness across distortion types.
While $\mathcal{M}\{2,3\}$ shows slightly reduced performance under clean and geometric conditions due to spatial mismatch between fixed embedding regions and dynamic tampering, it remains notably robust to video compression and frame-level distortions.
These findings illustrate how dimensional expansion facilitates localization functionalities that significantly surpass those achievable by one-dimensional baselines.

\subsection{High-to-Low Mapping: \boldmath$\mathcal{M}\{d_e>d_d\}$}

\noindent\textbf{Necessity of Decoupled Extraction.}
Figure~\ref{fig:channel_compare} analyzes the effect of multi-channel spatiotemporal payloads.
In $\mathcal{M}\{3,3\}$, performance drops significantly in higher channel dimensions. This behavior reflects that weakened cross-frame coherence restricts the model's capacity to resolve the full video volume.
In contrast, decoupling extraction into frame-wise prediction under $\mathcal{M}\{3,2\}$ restores localization performance, demonstrating the necessity of high-to-low dimensional projection in this regime.

\noindent\textbf{Impact of Multi-channel Mask Encoding.}
With four-channel mask encoding, the model achieves a PSNR of 39.44 dB and an SSIM of 0.9807. Although these values are slightly lower than those obtained with the single-channel encoding of DiMap-V, they remain superior to most baseline methods, indicating that the impact of multi-channel encoding on imperceptibility is negligible. We further evaluate the impact of the four-channel mask setting on extraction robustness and tamper localization. As illustrated in Figure~\ref{fig:mc4_acc_miou}, \vname maintains strong local extraction robustness under most distortion types, with moderate degradation under severe geometric transformations and compression due to the increased complexity of predicting multi-channel masks under such distortions. Regarding tamper localization, performance decreases for extremely small mask ratios. Nevertheless, this trade-off is expected, as simultaneously achieving precise localization for minute spatial regions while encoding temporal indices through multi-channel representations presents an inherent challenge.
Finally, Fig.~\ref{fig:merge_multi} shows that multi-channel encoding preserves frame identity under temporal permutation, enabling temporal order localization beyond spatial tamper detection.

\section{Conclusion}
We presented \name, a dimension-aware framework that unifies deep watermarking by modeling embedding and extraction as mappings between payloads of different dimensionalities. By instantiating \name in the video domain as \vname, we demonstrate that diverse watermarking behaviors, including local embedding, spatiotemporal tamper localization, and temporal order recovery, emerge naturally from different dimensional configurations.
These capabilities are achieved without architectural modifications, highlighting representation dimensionality as a central factor governing watermarking functionality.

\section*{Impact Statement}
This work presents a dimension-aware framework for understanding deep watermarking by modeling embedding and extraction as mappings between payloads of different dimensionalities. The framework is analytical and does not introduce new attack capabilities. Instead, it provides a unified perspective for interpreting existing watermarking methods and clarifying how different functionalities arise from representational choices.

When instantiated in video watermarking, the framework shows that capabilities such as spatial and spatiotemporal tamper localization and temporal order recovery can be achieved without architectural changes. These findings are relevant to applications in copyright protection, media authentication, and forensic analysis, and support more principled and interpretable watermarking system design.

\bibliography{main}
\bibliographystyle{icml2026}

\clearpage
\newpage
\appendix
\onecolumn
\section{More Details}
\subsection{Training Details.}
\label{sec: training details}
Under the single-channel mask setting, all videos are resized to 256×256 with 8 frames during training.
Under the multi-channel mask setting, we train on videos of 128×128 with 8 frames.
This reduced spatial resolution is adopted to improve training efficiency and memory usage. All models are trained on a single NVIDIA H100 GPU with a batch size of 8 for 200k steps.
We use the AdamW optimizer with a learning rate of \(2 \times 10^{-4}\), together with a cosine learning rate scheduler and 2,000 warm-up steps. Following MaskWM, we adopt a curriculum learning strategy from easy to hard.
During the first 1,000 steps, only full-one masks are used and no distortions are applied.
From 1,000 to 2,000 steps, all mask types are introduced.
After 2,000 steps, distortion layers are enabled. 
The encoder loss weight $\beta_{\text{enc}}$ is fixed to 1. The decoder loss weight $\beta_{\text{dec}}$ is initialized to 20 and linearly decayed to 0.2 over the first 10k steps.
The mask loss weight is set to 0.5. The JND module in the encoder is activated and optimized from 10k steps, with the scaling factor set to 1.

For VAE-based fine-tuning, VAE distortions are applied with 50\% probability, while the remaining 50\% use the original distortion types. The hyperparameters are set as follows: $\beta_{\text{enc}}=0.3$, learning rate \(2 \times 10^{-4}\), and the quality levels of \textit{Bmshj18} and \textit{Cheng20} are both set to 5.
\subsection{Distortion Details}
\subsubsection{training}
\label{sec:training distortion}
\noindent\textbf{Valuemetric Distortions.}
During training, valuemetric robustness is enhanced by randomly sampling from four common distortions: Gaussian blur, Gaussian noise, median filtering, and salt-and-pepper noise. The parameters are set as follows:

\begin{itemize}[leftmargin=20pt]

\item \textbf{Gaussian blur:} kernel size = 1, standard deviation = 5.
\item \textbf{Gaussian noise:} mean = 0, standard deviation = 0.1.
\item \textbf{Median filter:} kernel size = 5.
\item \textbf{Salt-and-pepper noise:} noise ratio = 0.1.
\end{itemize}

\noindent\textbf{Geometric Distortions.}
To improve geometric robustness, we randomly sample from three typical geometric transformations: rotation, perspective transformation, and horizontal flipping. The configurations are as follows:
\begin{itemize}[leftmargin=20pt]
\item \textbf{Rotation:} angle sampled from $[-90^\circ, 90^\circ]$.
\item \textbf{Perspective:} distortion scale sampled from $[0.1, 0.5]$.
\item \textbf{Horizontal flip:} no parameters.
\end{itemize}

\noindent\textbf{Video Compression.}
We additionally incorporate an H.264-like compression distortion during training.
Specifically, we adopt the distortion layer proposed in REVMark~\cite{zhang2023novel}, which simulates both
intra-frame and inter-frame compression effects.
During training, the intra-frame compression strength is randomly sampled from
$[1.5, 5]$, while the inter-frame compression strength is randomly sampled from
$[5, 8]$.

\noindent\textbf{Frame-level Distortions.}
To improve robustness to frame-related perturbations, we further apply the following distortions during training:
\begin{itemize}[leftmargin=20pt]
\item \textbf{Frame shuffling:} randomly permute the frame order, with no constraint on the number of affected frames; in the most severe case, all frames are fully shuffled.
\item \textbf{Frame replacement:} randomly replace one frame with an all-white frame (simulating frame substitution).
\item \textbf{Frame dropping:} randomly drop one frame and append a new frame at the end (simulating frame loss).
\item \textbf{Frame insertion:} randomly insert an all-white frame and remove the last frame (simulating insertion of an unrelated frame).
\end{itemize}

\subsubsection{evaluation}
\label{sec: test distortion}
\noindent\textbf{Valuemetric Distortions.}
We apply four valuemetric distortions and evaluate robustness using the following parameter settings:
\begin{itemize}[leftmargin=20pt]
\item \textbf{JPEG compression:} quality factor = 60.
\item \textbf{Gaussian blur:} kernel size = 1, standard deviation = 3.
\item \textbf{Gaussian noise:} mean = 0, standard deviation = 0.05.
\item \textbf{Median filter:} kernel size = 3.
\item \textbf{Salt-and-pepper noise:} noise ratio = 0.05.
\end{itemize}

\noindent\textbf{Geometric Distortions.}
We apply three geometric distortions and evaluate robustness using the following configurations:
\begin{itemize}[leftmargin=20pt]
\item \textbf{Rotation:} angle sampled from $[-30^\circ, 30^\circ]$.
\item \textbf{Perspective:} distortion scale sampled from $[0.1, 0.3]$.
\item \textbf{Horizontal flip:} no parameters.
\end{itemize}

\noindent\textbf{Video Compression.}
We apply the built-in H.264 compression provided by the
\texttt{torchvision.io} library when saving videos. Videos are encoded using the H.264 codec with a constant rate factor (CRF),
where we evaluate two compression levels with CRF = 20 and CRF = 25.
Lower CRF values correspond to higher visual quality and weaker compression,
while higher CRF values indicate stronger compression.

\noindent\textbf{Frame-level Distortions.}
We further evaluate robustness against frame-related perturbations using the same
set of operations as in training, including:
\begin{itemize}[leftmargin=20pt]
\item \textbf{Frame shuffling:} randomly permuting the frame order, with no constraint on the number of affected frames.
\item \textbf{Frame replacement:} randomly replacing one frame with an all-white frame.
\item \textbf{Frame dropping:} randomly dropping one frame and appending a new all-white frame at the end.
\item \textbf{Frame insertion:} randomly inserting an all-white frame and removing the last frame.
\end{itemize}

\section{Training}
\label{sec:training_pipeline}
\subsection{Unified Watermark Pipeline}
\label{sec:embedding_pipeline}

The encoder first processes the concatenated input and produces an intermediate encoded video $\mathbf{V}_{\text{enc}}$.
To enhance perceptual quality, we apply a Just-Noticeable Difference (JND) module that adaptively modulates the embedding signal according to human visual sensitivity, resulting in the final watermarked video $\mathbf{V}_{\text{wm}}$.
\begin{equation}
\mathbf{V}_{\text{wm}}=\mathbf{V}_{\text{orig}} +\mu \times \text{JND}(\mathbf{V}_{\text{orig}})\times(\mathbf{V}_{\text{enc}}-\mathbf{V}_{\text{orig}})
\end{equation}
To endow the model with mask-controlled embedding and extraction capabilities, as well as precise localization of manipulated regions, we adopt the MaskWM fusion strategy:
\begin{equation}
\mathbf{V}_{\text{fuse}} = \mathbf{V}_{\text{wm}} \odot \mathbf{M}^{(3)} + \mathbf{V}_{\text{orig}} \odot (1 - \mathbf{M}^{(3)}),
\end{equation}
Masked regions preserve the embedded watermark, while unmasked regions are replaced by the original video content.
The fused video $\mathbf{V}_{\text{fuse}}$ is then passed through a noise layer $\mathcal{A}$, which models a stochastic attack channel by randomly sampling transformations from a predefined distortion pool (see Appendix~\ref{sec: training details} for details). The resulting video $\mathbf{V}_{\text{fusion}}'$ captures a wide range of appearance, geometric, and temporal degradations commonly encountered in practical scenarios. 
\begin{equation}
\mathbf{V}_{\text{fusion}}' = \mathcal{A}(\mathbf{V}_{\text{fuse}}).
\end{equation}
Finally, the masked extraction input is constructed as
\begin{equation}
\mathbf{V}_{\text{mask}} = \mathbf{V}_{\text{fusion}}' \odot \mathbf{M},
\end{equation}
which isolates watermark-bearing regions for subsequent watermark recovery $\mathbf{W}_{\mathrm{pd}}$ in decoder, while the fused video $\mathbf{V}_{\text{fusion}}'$ is simultaneously used by the mask prediction network to infer the embedded mask $\mathbf{M}^{(3)}_\text{pd}$.

\subsection{Network Architectures}
\label{sec:architectures}

\noindent\textbf{Message Translator.}
A linear layer first maps $\mathbf{p}^{(1)}$ to a latent tensor of shape
$(1,\, L,\, L,\, L \times (H/T))$,
which is then resized via trilinear interpolation to $(1,\, T,\, H,\, W)$.
A lightweight 3D CNN composed of multiple Conv--Norm--ReLU (CNR) blocks transforms this tensor into the final message representation $\mathbf{T}_{\text{msg}} \in \mathbb{R}^{C_{tp} \times T \times H \times W}$.

\noindent\textbf{Encoder and Decoder.}
The encoder and decoder are adapted from MaskWM by extending all 2D convolutional modules to 3D convolutions, enabling direct operation on spatiotemporal video volumes. The encoder embeds the message and mask into the input video to produce a watermarked video, while the decoder recovers the 1D binary payload from $\mathbf{V}_{\text{mask}}$.

\noindent\textbf{Mask Prediction Networks.}
To support both spatial and spatiotemporal payload extraction, we employ two mask prediction networks.
For 2D spatial payload prediction in $\mathcal{M}\{3,2\}$, we adopt the same U$^2$-Net architecture as MaskWM, which effectively captures fine-grained spatial structures.
For 3D spatiotemporal payloads prediction in $\mathcal{M}\{3,3\}$, $\mathcal{M}\{1,3\}$ and $\mathcal{M}\{2,3\}$, we extend U$^2$-Net by replacing all 2D convolutional modules with 3D counterparts, preserving its hierarchical multi-scale design while enabling temporal modeling.

\noindent\textbf{Training Objectives.} We follow the loss design of MaskWM. The overall training objective is defined as
\begin{equation}
\mathcal{L}_{\mathrm{total}} =
\beta_{\mathrm{enc}} \mathcal{L}_{\mathrm{enc}}
+
\beta_{\mathrm{dec}} \mathcal{L}_{\mathrm{dec}},
\end{equation}
where \(\beta_{\mathrm{enc}}\) and \(\beta_{\mathrm{dec}}\) balance the encoder and decoder losses, respectively. The encoder and decoder losses are defined as
\begin{align}
\mathcal{L}_{\mathrm{enc}}
&= \mathcal{L}_{\mathrm{MSE}}\!\left(
\mathbf{V}_{\mathrm{wm}},\;
\mathbf{V}_{\mathrm{orig}}
\right), \\
\mathcal{L}_{\mathrm{dec}}
&= \mathcal{L}_{\mathrm{MSE}}\!\left(
\mathbf{W}_{\mathrm{pd}},\;
\mathbf{W}
\right)
+ \alpha \,\mathcal{L}_{\mathrm{MSE}}\!\left(
\mathbf{M}^{(3)}_{\mathrm{pd}},\;
\mathbf{M}^{(3)}
\right),
\end{align}
where $\alpha$ controls the weight of the mask loss.

\section{Mask Shifting strategy}
\label{sec: mask shift}
Algorithm~\ref{alg:seq_gen} generates a spatiotemporal mask sequence by propagating a single 2D mask across time using randomized spatial shifts. At each frame, a displacement direction and magnitude are sampled, and the current mask is shifted accordingly. The shifted mask is accepted if it remains non-empty after boundary handling, ensuring valid mask propagation. Repeating this process yields a temporally coherent mask sequence that emulates object-tracking–like behavior while avoiding degenerate empty masks.

\begin{algorithm}[tb]
\caption{Spatiotemporal Mask Sequence Generation}
\label{alg:seq_gen}
\begin{algorithmic}[1]
\STATE {\bfseries Input:} Initial 2D mask $\mathbf{M}_0$, Sequence length $T$, Max movement $\delta_{\max}$
\STATE {\bfseries Output:} Spatiotemporal mask sequence $\mathcal{S} = \{\mathbf{M}_0, \dots, \mathbf{M}_{T-1}\}$

\STATE Initialize sequence $\mathcal{S} \leftarrow \{\mathbf{M}_0\}$, set current mask $\mathbf{M}_{\text{curr}} \leftarrow \mathbf{M}_0$

\FOR{$t = 1$ {\bfseries to} $T-1$}
\STATE \textcolor{gray}{\textit{\# Initialize and shuffle candidate directions $\mathcal{D}$}}
\STATE $\mathcal{D} \leftarrow \{(x,y) \in \{-1, 0, 1\}^2 \mid (x,y) \neq (0, 0)\}$
\STATE Randomly permute the order of $\mathcal{D}$

\FOR{\textbf{each} direction $\mathbf{d} \in \mathcal{D}$}
\STATE \textcolor{gray}{\textit{\# Sample magnitude and compute displacement}}
\STATE Sample step size $k \sim \mathcal{U}_{\text{int}}(0, \delta_{\max})$
\STATE Compute shift vector $(\Delta x, \Delta y) \leftarrow k \cdot \mathbf{d}$

\STATE \textcolor{gray}{\textit{\# Apply spatial shift and check boundary validity}}
\STATE $\mathbf{M}' \leftarrow \textsc{Shift}(\mathbf{M}_{\text{curr}}, \Delta x, \Delta y)$

\STATE \textcolor{gray}{\textit{\# Accept if the mask is not empty (contains active pixels)}}
\IF{$\sum \mathbf{M}' > 0$}
   \STATE $\mathbf{M}_{\text{curr}} \leftarrow \mathbf{M}'$
   \STATE $\mathcal{S} \leftarrow \mathcal{S} \cup \{\mathbf{M}_{\text{curr}}\}$
   \STATE \textbf{break}
\ENDIF
\ENDFOR

\ENDFOR

\RETURN $\,\mathcal{S}$
\STATE \hrulefill
\STATE \textbf{Function} $\textsc{Shift}(\mathbf{X}, \Delta x, \Delta y)$:
\STATE \textcolor{gray}{\textit{\# Initialize empty tensor}}
\STATE Initialize $\mathbf{X}_{\text{new}} \leftarrow \mathbf{0}$
\STATE Let $\Omega = [0, H-1] \times [0, W-1]$ be the spatial bounds

\STATE \textcolor{gray}{\textit{\# Get all active indices}}
\STATE Let $\mathcal{I} = \{(c, y, x) \mid \mathbf{X}_{c,y,x} \neq 0\}$ be the set of active indices

\STATE \textcolor{gray}{\textit{\# Parallel coordinate shift}}
\STATE Let $\mathcal{I}' = \{(c, y+\Delta y, x+\Delta x) \mid (c, y, x) \in \mathcal{I}\}$

\STATE \textcolor{gray}{\textit{\# Filter boundary and assign values}}
\STATE Identify valid subset $\mathcal{I}'_{\text{valid}} = \{(c, y', x') \in \mathcal{I}' \mid (y', x') \in \Omega\}$
\STATE Map values from $\mathbf{X}$ to $\mathbf{X}_{\text{new}}$ at indices $\mathcal{I}'_{\text{valid}}$

\RETURN $\,\mathbf{X}_{\text{new}}$
\end{algorithmic}
\end{algorithm}

\section{Further Analysis}
\subsection{Compression Robustness Enhancement}
\label{sec: compression improve}
We observe that increasing payload capacity naturally degrades robustness to compression. To mitigate this effect, we explore a targeted fine-tuning strategy that incorporates Variational Autoencoder (VAE)-simulated distortions into the training process. As shown in Table~\ref{tab:finetune_ablation}, this approach yields a substantial 6.18\% improvement in compression robustness while introducing only minor trade-offs under other distortions.
\begin{table*}[!h]
\centering
\caption{
Effect of compression-oriented fine-tuning on robustness and imperceptibility. We report the trade-offs between improved compression robustness and performance under other distortions.
Both strategies achieve 100\% extraction accuracy under no distortion.
}
\vspace{-0.8em}
\label{tab:finetune_ablation}
\setlength{\tabcolsep}{3pt}
\begin{center}
\begin{small}
\begin{sc}
\begin{tabular}{lccccccc}
\toprule
& \multirow{2}{*}{\textbf{PSNR $\uparrow$} } & \multirow{2}{*}{\textbf{SSIM $\uparrow$}} & \multirow{2}{*}{\textbf{No Distortion}}&\multicolumn{2}{c}{\textbf{Common Distortions $\uparrow$}}& \multicolumn{2}{c}{\textbf{Video Distortions $\uparrow$}} \\
\cmidrule(lr){5-6} \cmidrule(lr){7-8} 
&  &  &  &  Valuemetric & Geometric & Frame-level &  Compression \\
 \midrule
Before& 42.36 & 0.9855 & 100 & 100 & 99.73 & 100 & 90.11 \\
After & 39.62 \loss{-2.90} & 0.9765 \loss{-0.0101}&100 \steady{+0.00}& 100 \steady{+0.00}& 98.95 \loss{-0.78}& 100 \steady{+0.00}& 96.29 \gain{+6.18} \\
\bottomrule
\end{tabular}
\end{sc}
\end{small}
\end{center}
\end{table*}

\subsection{Efficiency}
\label{sec:efficiency}
Table~\ref{tab:speed} reports the computational efficiency in terms of Frames Per Second (FPS). \vname achieves an embedding speed of 114.66 FPS and an extraction speed of 228.57 FPS, exceeding all baseline methods. The high extraction throughput underscores the applicability of our \vname to real-time video watermarking tasks.
\begin{table*}[!h]
\caption{Comparison of computational efficiency with baseline methods in terms of frames per second (FPS).}
\vspace{-0.8em}
\label{tab:speed}
\centering
\setlength{\tabcolsep}{2pt}
\begin{center}
\begin{small}
\begin{sc}
\begin{tabular}{lccccccccc}
\toprule
\multirow{2}{*}{\textbf{Method}}  & \multicolumn{5}{c}{\textbf{Image Watermarking}} 
 & \multicolumn{4}{c}{\textbf{Video Watermarking}} \\
\cmidrule(lr){2-6} \cmidrule(lr){7-10}
& MaskWM & WAM & OmniGuard & Robust-Wide & TrustMark 
& VideoSeal & RivaGAN & REVMark & \textbf{Ours} \\
\midrule
Embed   
& 48.2793 & 38.4982 & 3.1480 & 59.2122 & 87.5891 
& 87.7029 & 2.2787 & 59.7755 & \textbf{114.6625} \\

Extract 
& 19.0688 & 28.0508 & 0.6189 & 18.1917 & 35.3368 
& 97.6562 & 2.4554 & 83.7395 & \textbf{228.5714} \\
\bottomrule
\end{tabular}
\end{sc}
\end{small}
\end{center}
\end{table*}

\section{More Visual Results}
\subsection{Global Watermarking}
\begin{figure*}[!t]
    \centering
    \includegraphics[width=\linewidth]{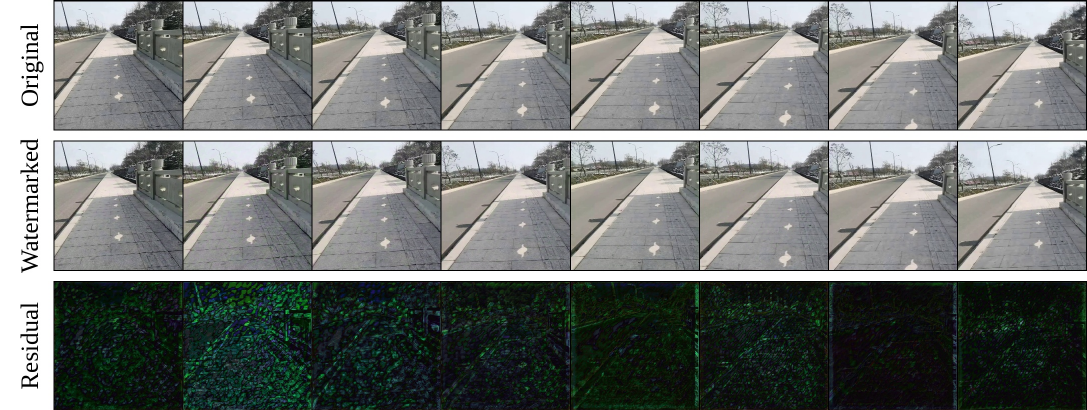}
    \caption{Visualization results of global watermark embedding using $\mathcal{M}\{1,3\}$. The residual images are magnified by 2×.}
    \label{fig:visual_1_3}
\end{figure*}

\begin{figure*}[!t]
    \centering
    \includegraphics[width=\linewidth]{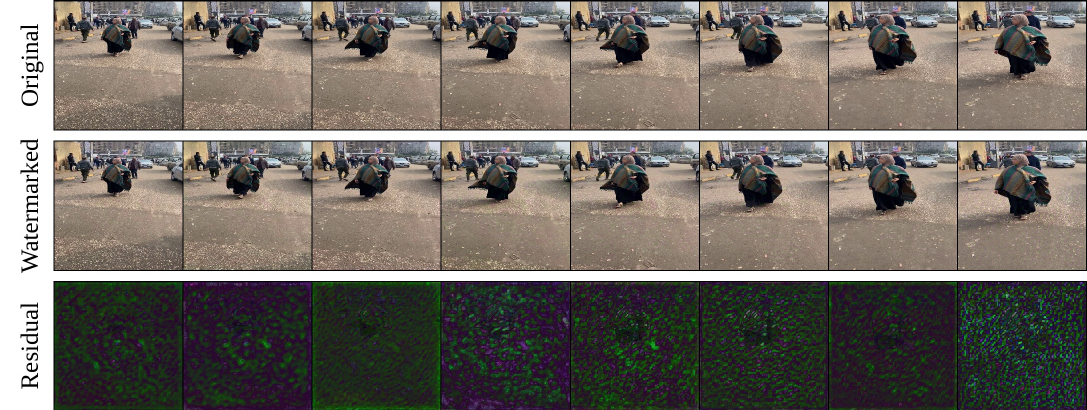}
    \caption{Visualization results of global watermark embedding using $\mathcal{M}\{2,3\}$. The residual images are magnified by 2×.}
    \label{fig:visual_2_3}
\end{figure*}

\begin{figure*}[!t]
    \centering
    \includegraphics[width=\linewidth]{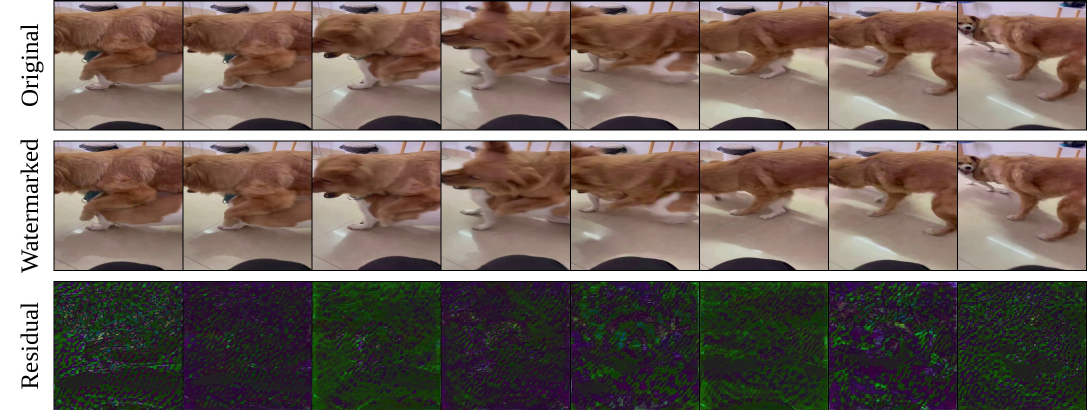}
    \caption{Visualization results of global watermark embedding using $\mathcal{M}\{3,3\}$. The residual images are magnified by 2×.}
    \label{fig:visual_3_3}
\end{figure*}

\begin{itemize}
    \item $\mathcal{M}\{1,3\}$: see Figure~\ref{fig:visual_1_3}.
    \item $\mathcal{M}\{2,3\}$: see Figure~\ref{fig:visual_2_3}.
    \item $\mathcal{M}\{3,3\}$: see Figure~\ref{fig:visual_3_3}.
\end{itemize}

\subsection{Local Watermarking}
\begin{figure*}
    \centering
    \includegraphics[width=\linewidth]{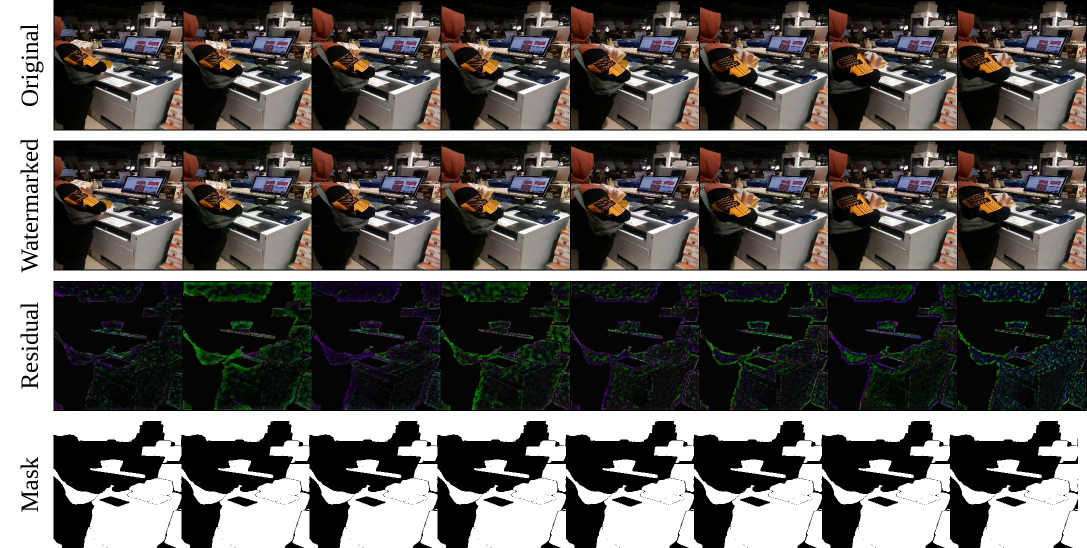}
    \caption{Visualization results of local watermark embedding using $\mathcal{M}\{2,3\}$. The residual images are magnified by 2×.}
    \label{fig:localvisual_2_3}
\end{figure*}

\begin{figure*}
    \centering
    \includegraphics[width=\linewidth]{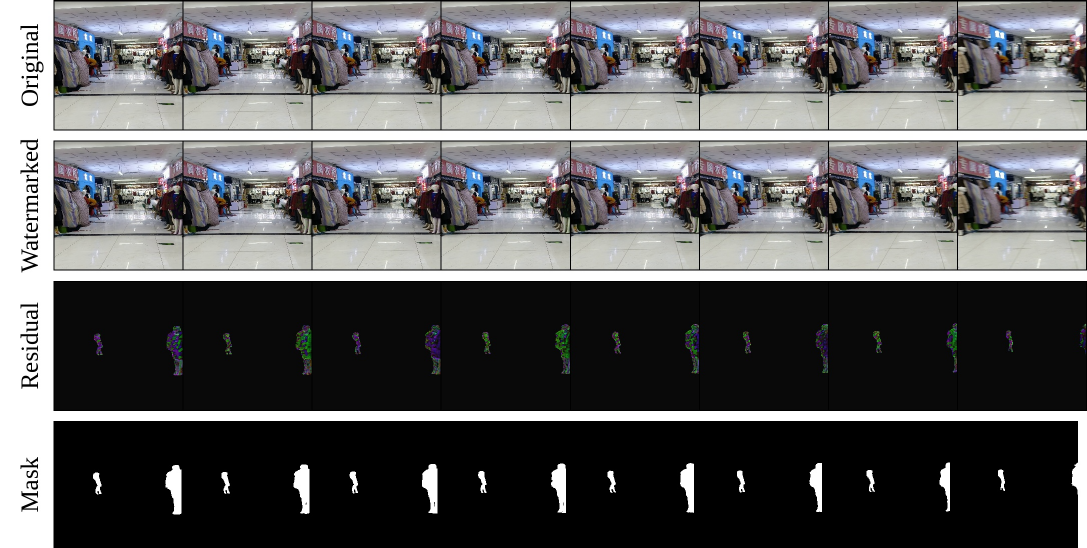}
    \caption{Visualization results of local watermark embedding using $\mathcal{M}\{3,3\}$. The residual images are magnified by 2×.}
    \label{fig:localvisual_3_3}
\end{figure*}

\begin{itemize}
    \item $\mathcal{M}\{2,3\}$: see Figure~\ref{fig:localvisual_2_3}.
    \item $\mathcal{M}\{3,3\}$: see Figure~\ref{fig:localvisual_3_3}.
\end{itemize}

\subsection{Visualization Results of Localization}
\begin{figure*}
    \centering
    \includegraphics[width=\linewidth]{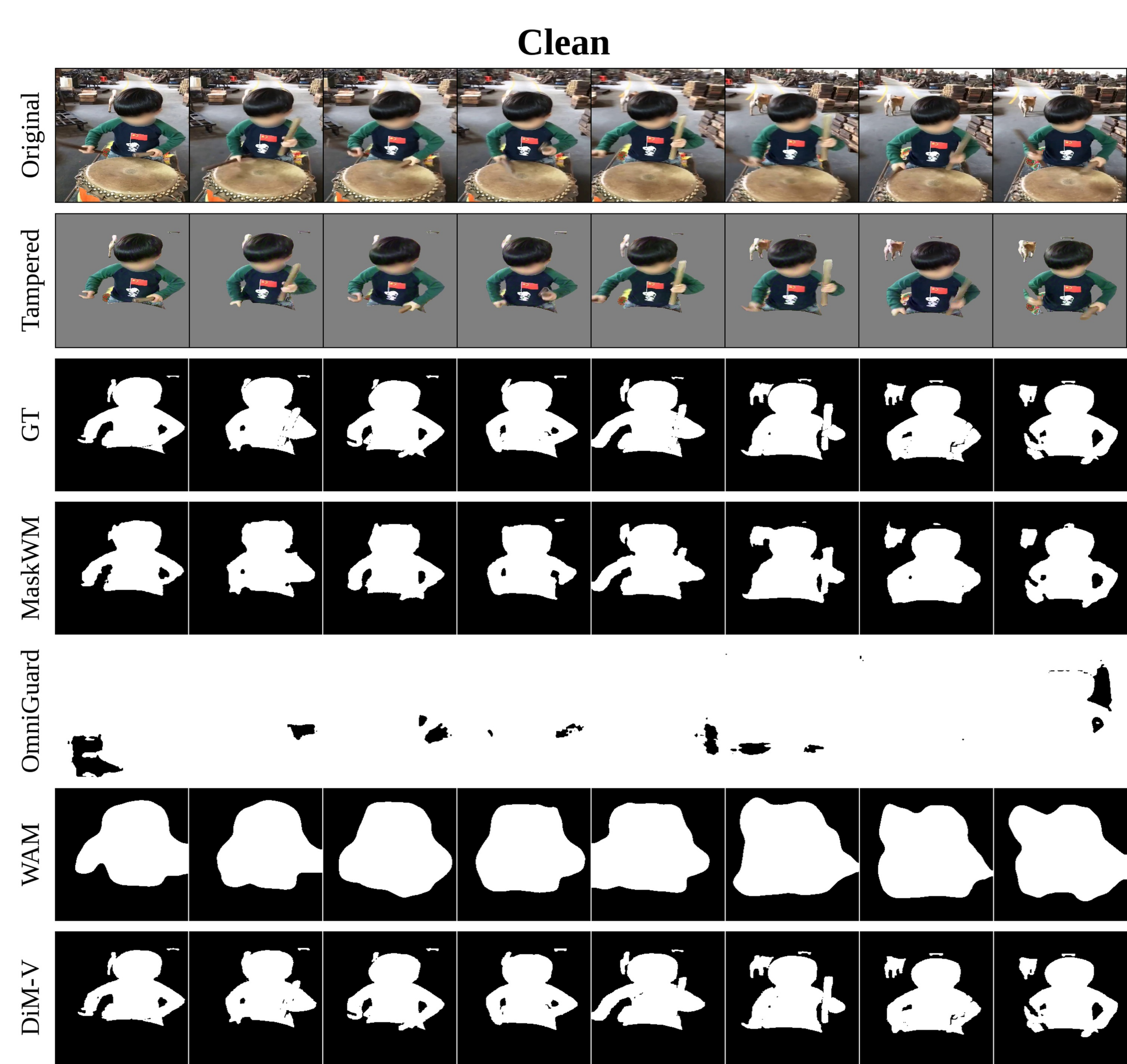}
    \caption{Visualization results of watermark localization using different methods.}
    \label{fig:localize_clean}
\end{figure*}

\begin{figure*}
    \centering
    \includegraphics[width=\linewidth]{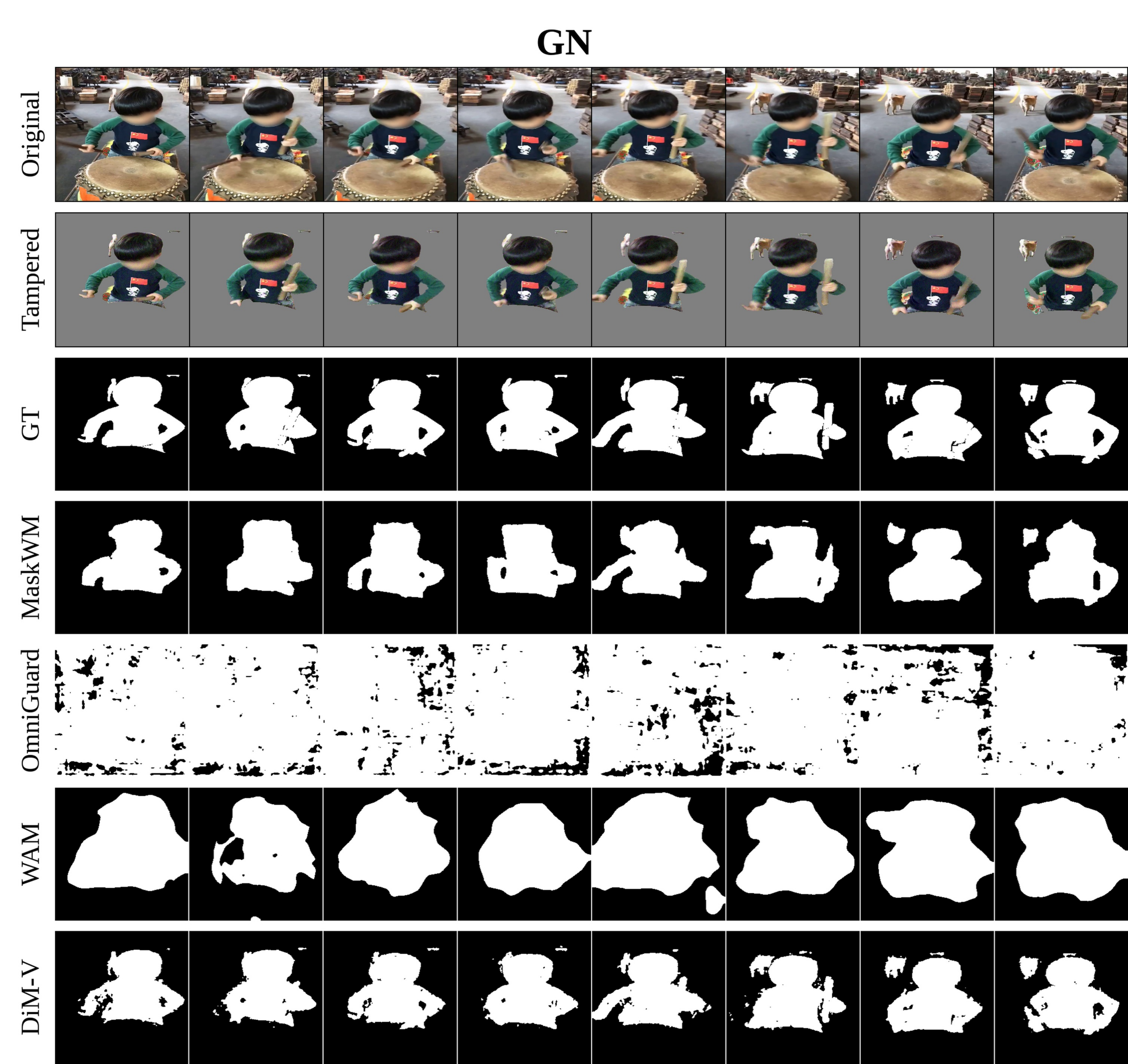}
    \caption{Visualization results of watermark localization using different methods under Gaussian noise.}
    \label{fig:localize_GN}
\end{figure*}

\begin{figure*}
    \centering
    \includegraphics[width=\linewidth]{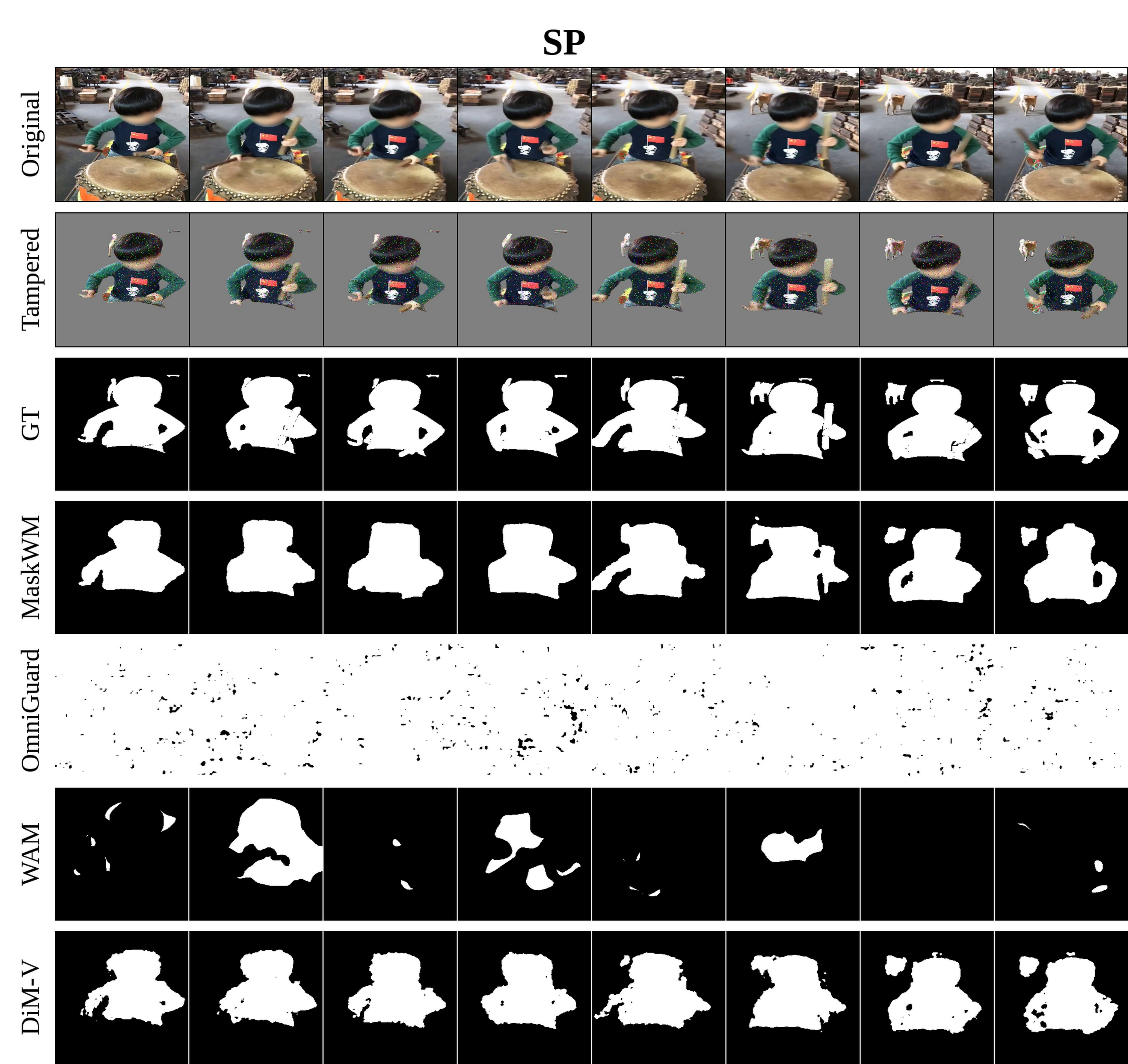}
    \caption{Visualization results of watermark localization using different methods under salt-and-pepper noise.}
    \label{fig:localize_SP}
\end{figure*}

\begin{figure*}
    \centering
    \includegraphics[width=\linewidth]{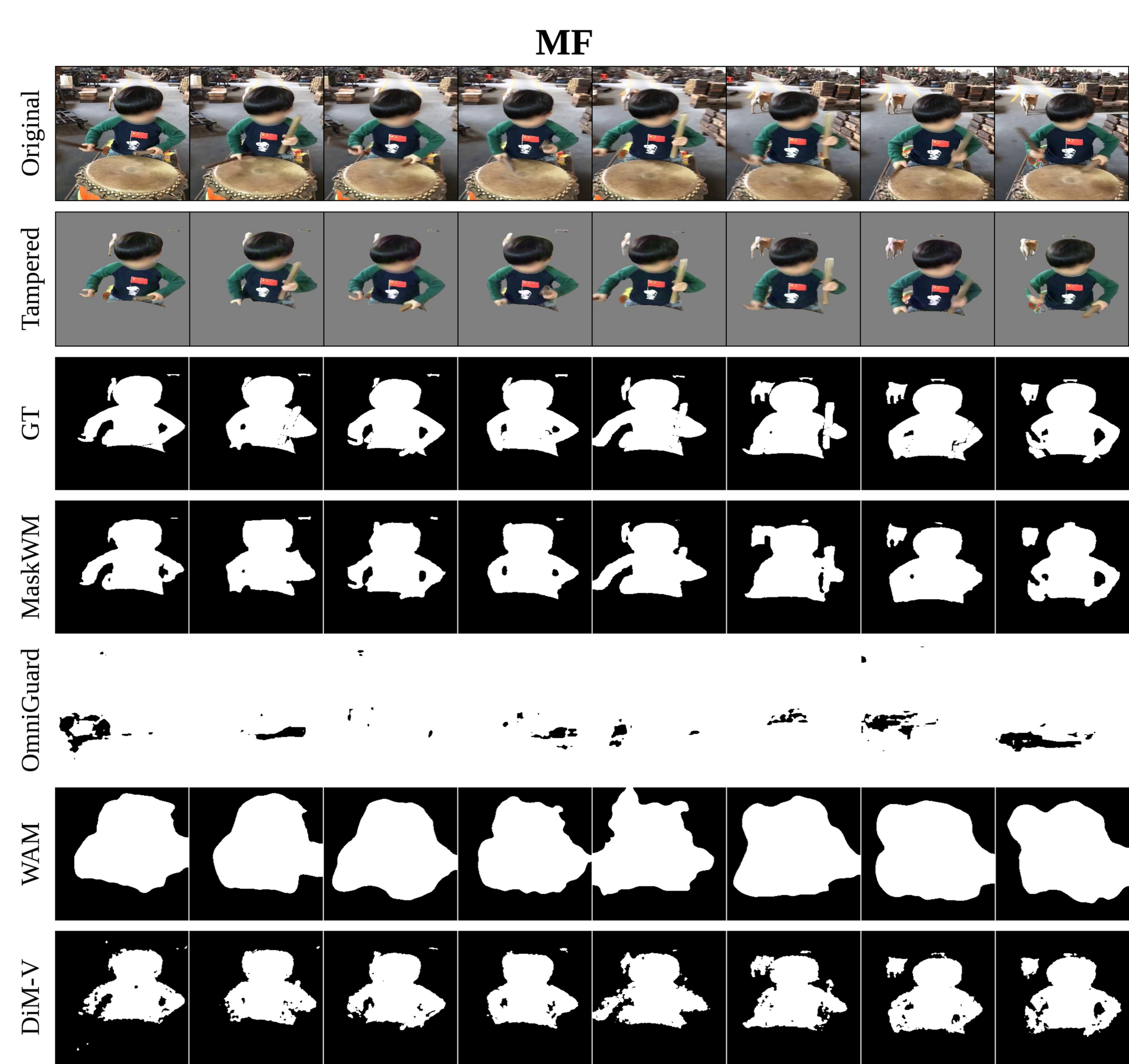}
    \caption{Visualization results of watermark localization using different methods under median filtering.}
    \label{fig:localize_MF}
\end{figure*}

\begin{figure*}
    \centering
    \includegraphics[width=\linewidth]{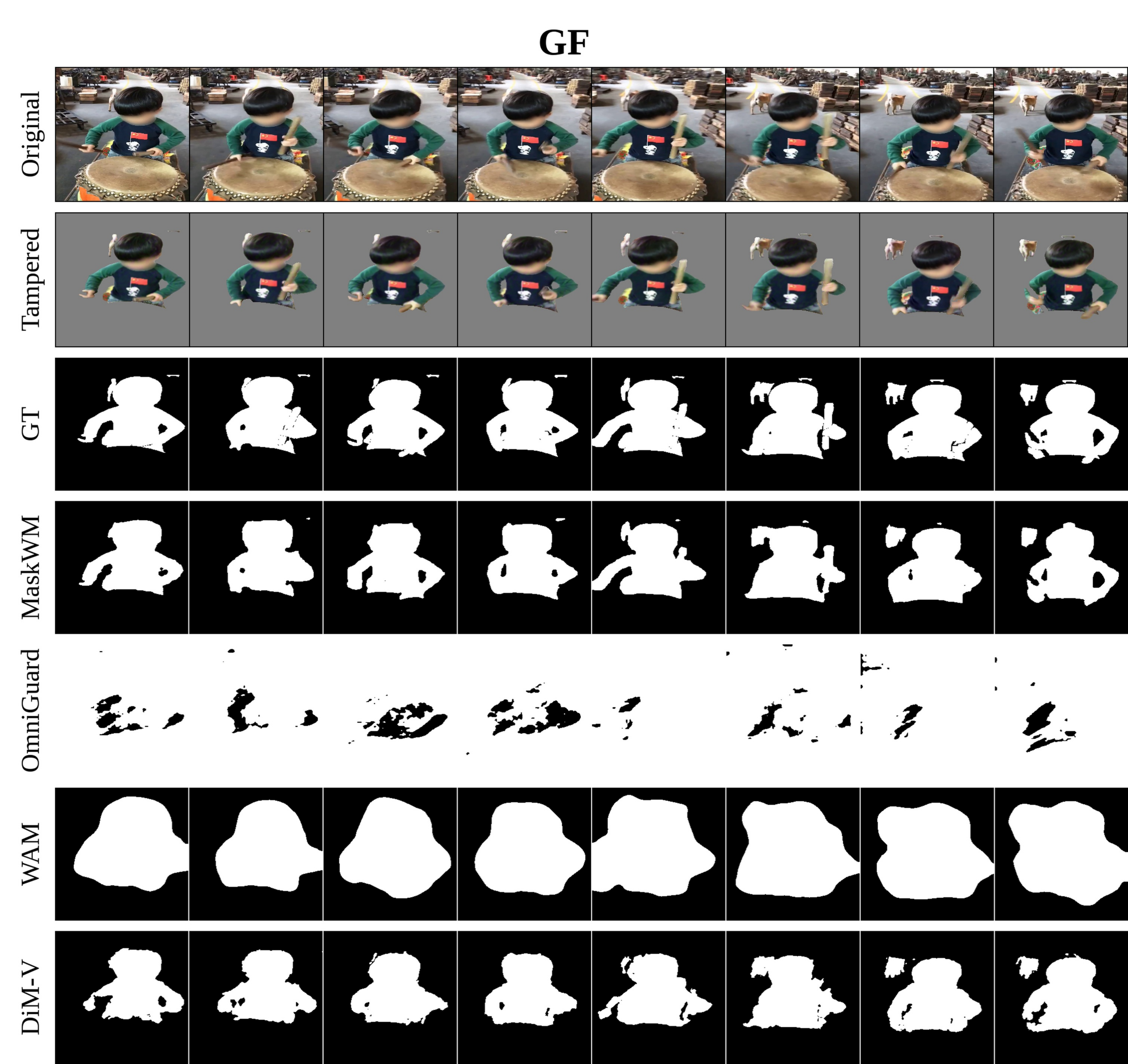}
    \caption{Visualization results of watermark localization using different methods under Gaussian blur.}
    \label{fig:localize_GF}
\end{figure*}

\begin{figure*}
    \centering
    \includegraphics[width=\linewidth]{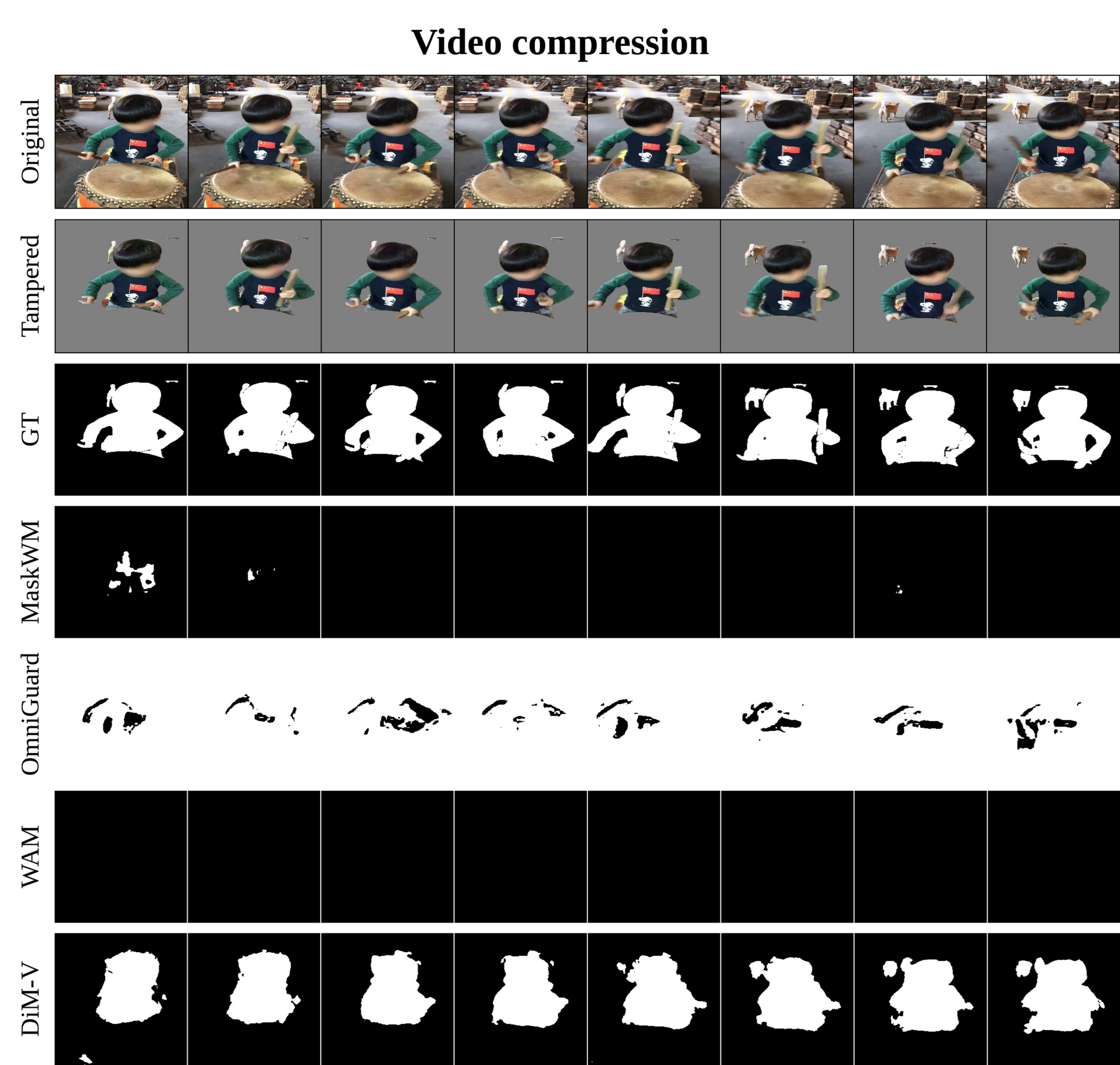}
    \caption{Visualization results of watermark localization using different methods under video compression.}
    \label{fig:localize_compression}
\end{figure*}

\begin{figure*}
    \centering
    \includegraphics[width=\linewidth]{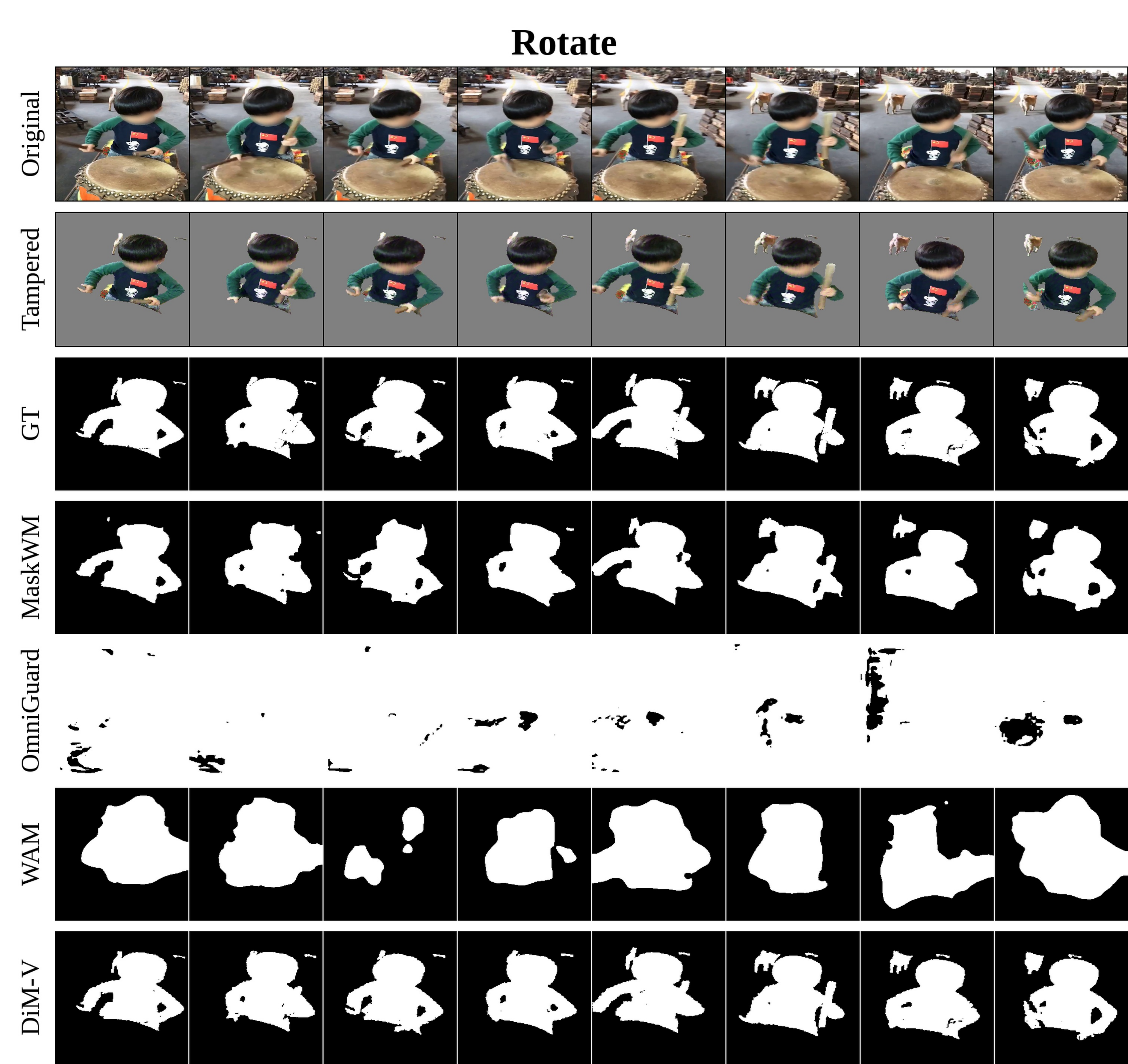}
    \caption{Visualization results of watermark localization using different methods under rotation.}
    \label{fig:localize_Rotate}
\end{figure*}

\begin{figure*}
    \centering
    \includegraphics[width=\linewidth]{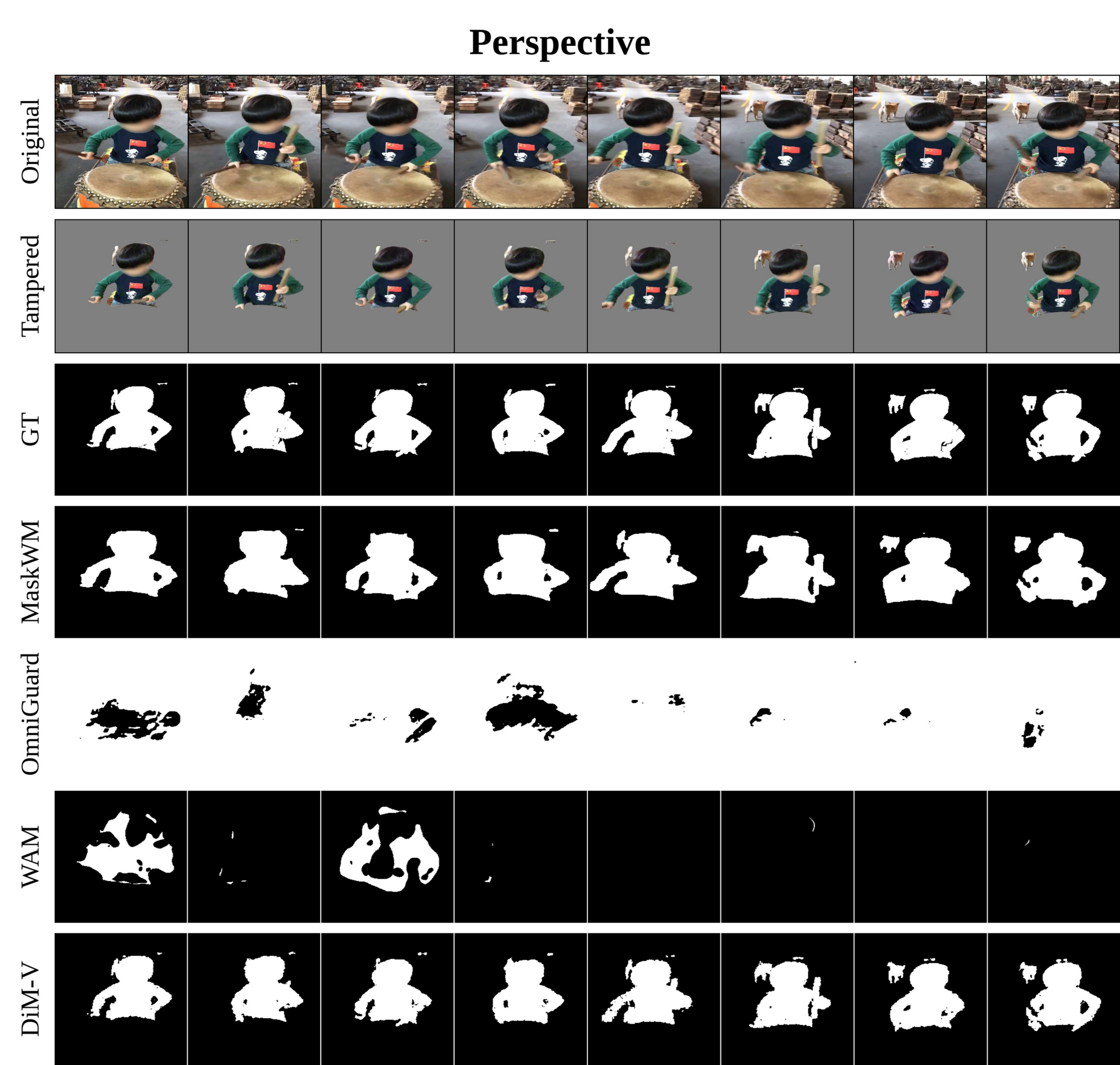}
    \caption{Visualization results of watermark localization using different methods under perspective transformation.}
    \label{fig:localize_perspective}
\end{figure*}

\begin{figure*}
    \centering
    \includegraphics[width=\linewidth]{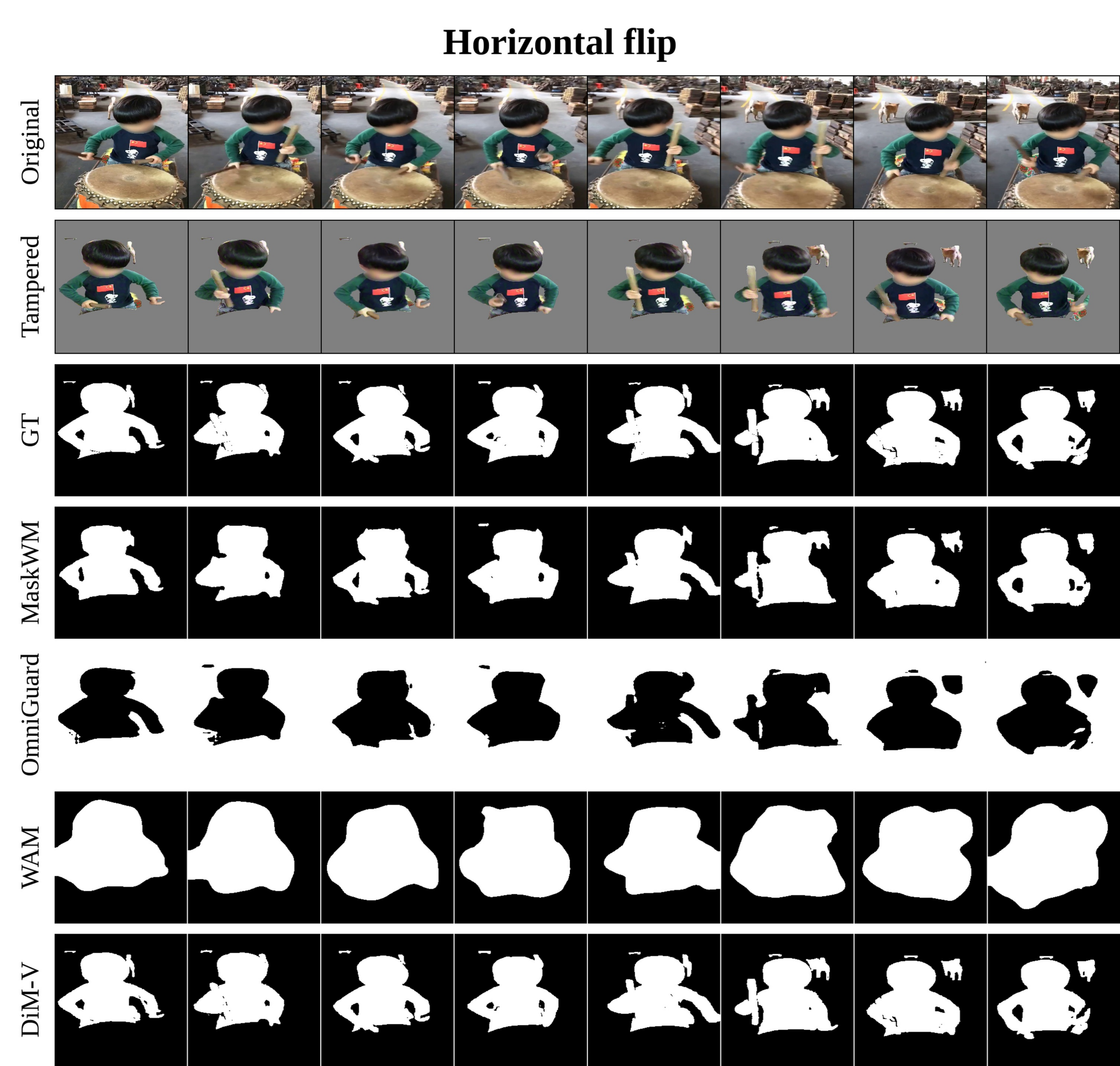}
    \caption{Visualization results of watermark localization using different methods under horizontal flipping.}
    \label{fig:localize_hf}
\end{figure*}

\begin{figure*}
    \centering
    \includegraphics[width=\linewidth]{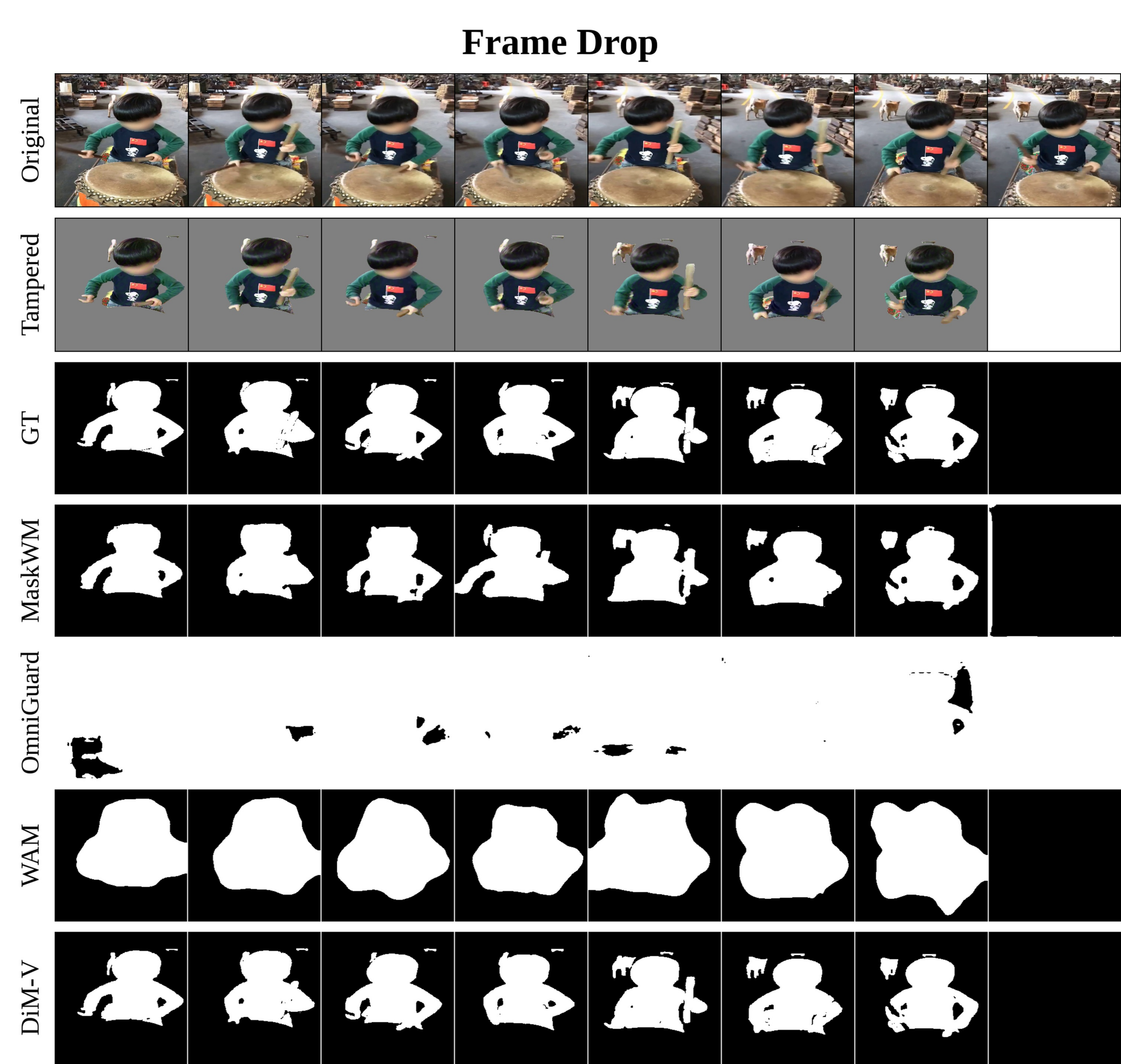}
    \caption{Visualization results of watermark localization using different methods under frame dropping.}
    \label{fig:localize_droppad}
\end{figure*}

\begin{figure*}
    \centering
    \includegraphics[width=\linewidth]{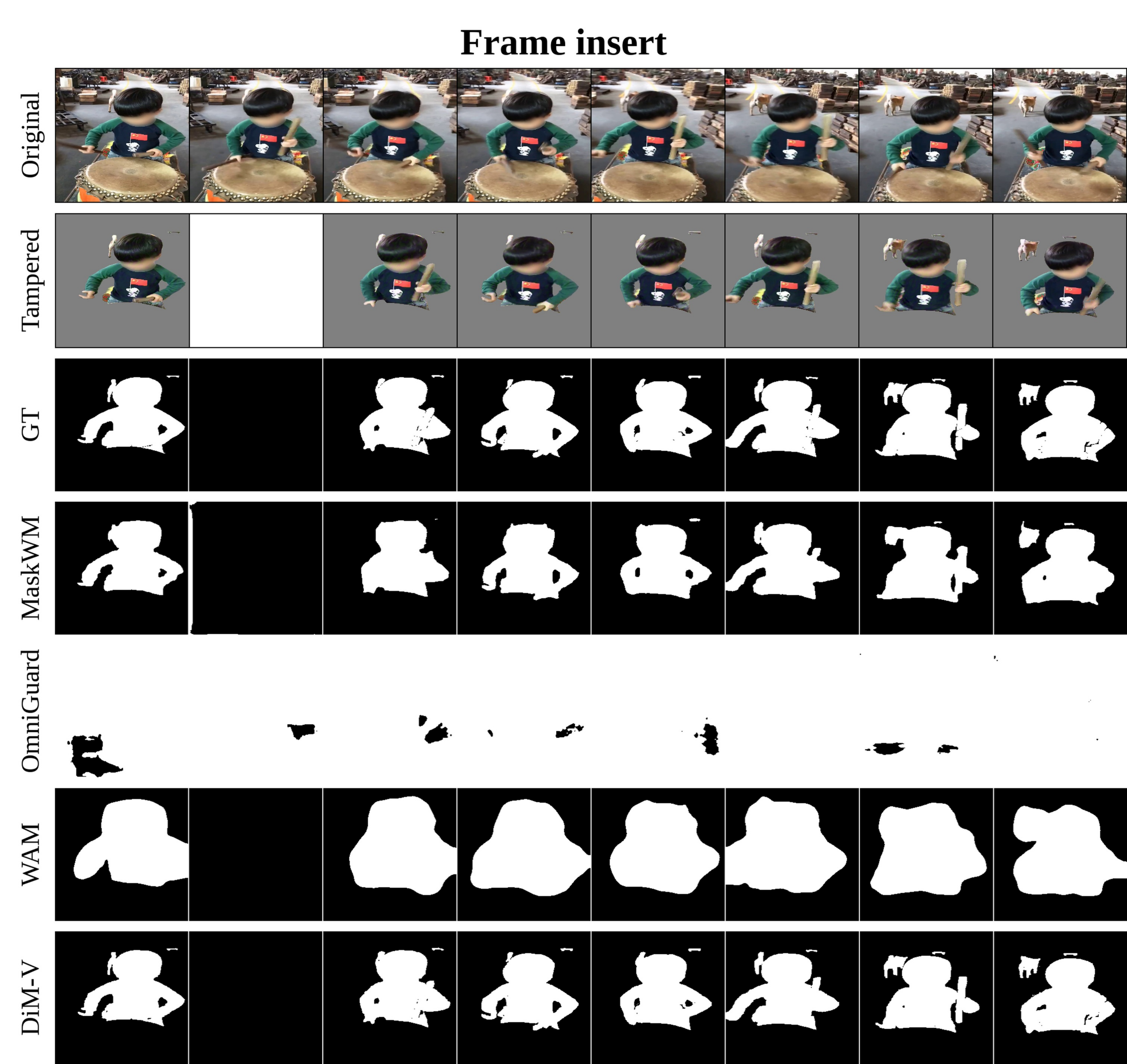}
    \caption{Visualization results of watermark localization using different methods under frame insertion.}
    \label{fig:localize_insert}
\end{figure*}

\begin{itemize}
    \item No distortion: see Figure~\ref{fig:localize_clean}.
    \item Gaussian noise: see Figure~\ref{fig:localize_GN}.
    \item Salt-and-pepper noise: see Figure~\ref{fig:localize_SP}.
    \item Median filtering: see Figure~\ref{fig:localize_MF}.
    \item Gaussian blur: see Figure~\ref{fig:localize_GF}.
    \item Video compression: see Figure~\ref{fig:localize_compression}.
    \item Rotation: see Figure~\ref{fig:localize_Rotate}.
    \item Perspective transformation: see Figure~\ref{fig:localize_perspective}.
    \item Horizontal flipping: see Figure~\ref{fig:localize_hf}.
    \item Frame dropping: see Figure~\ref{fig:localize_droppad}.
    \item Frame insertion: see Figure~\ref{fig:localize_insert}.
\end{itemize}

\subsection{Comparison of \boldmath$\mathcal{M}\{3,3\}$ and \boldmath$\mathcal{M}\{3,2\}$ in Multi-Channel Mask Prediction}
\begin{figure*}[!h]
    \centering
    \includegraphics[width=\linewidth]{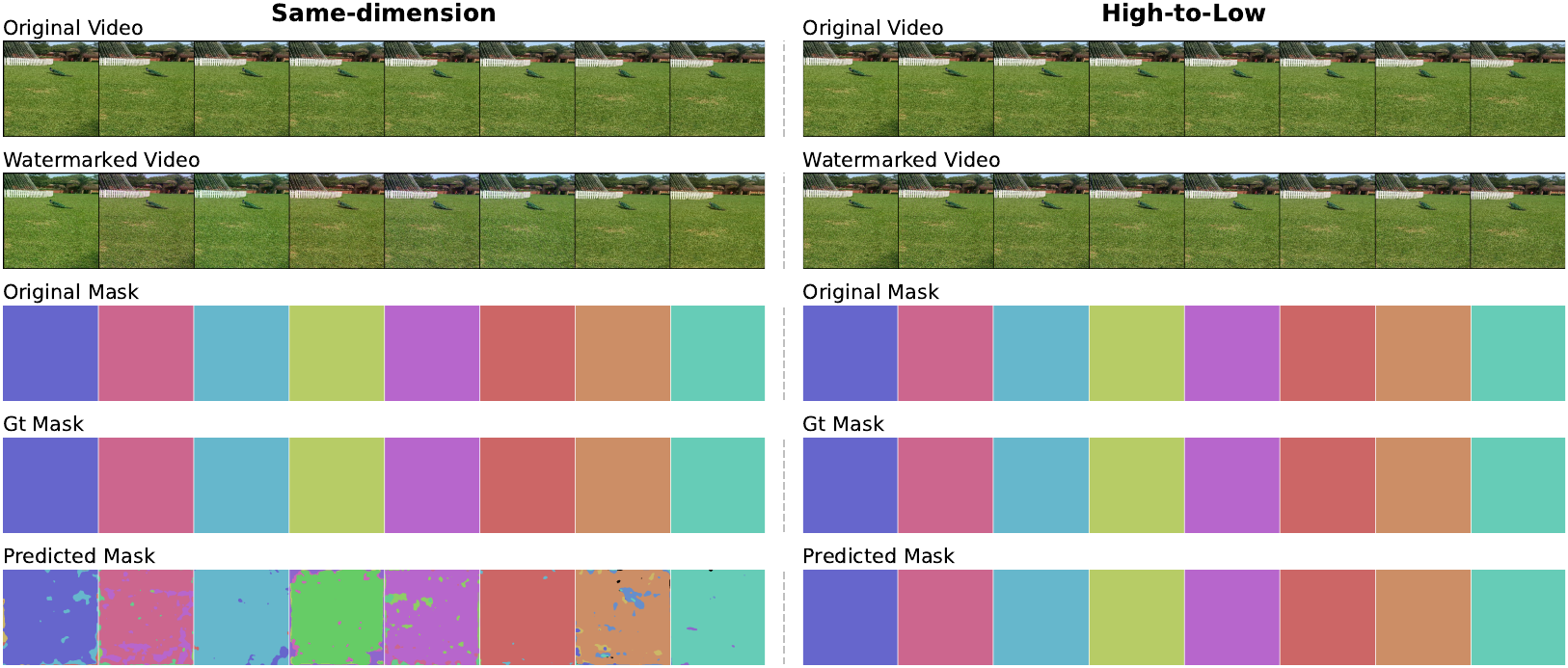}
    \caption{Visualization of predicted multi-channel masks for $\mathcal{M}\{3,3\}$ and $\mathcal{M}\{3,2\}$. Different mask encodings are visualized using distinct colors.}
    \label{fig:multi_clean}
\end{figure*}

\begin{figure*}[!h]
    \centering
    \includegraphics[width=\linewidth]{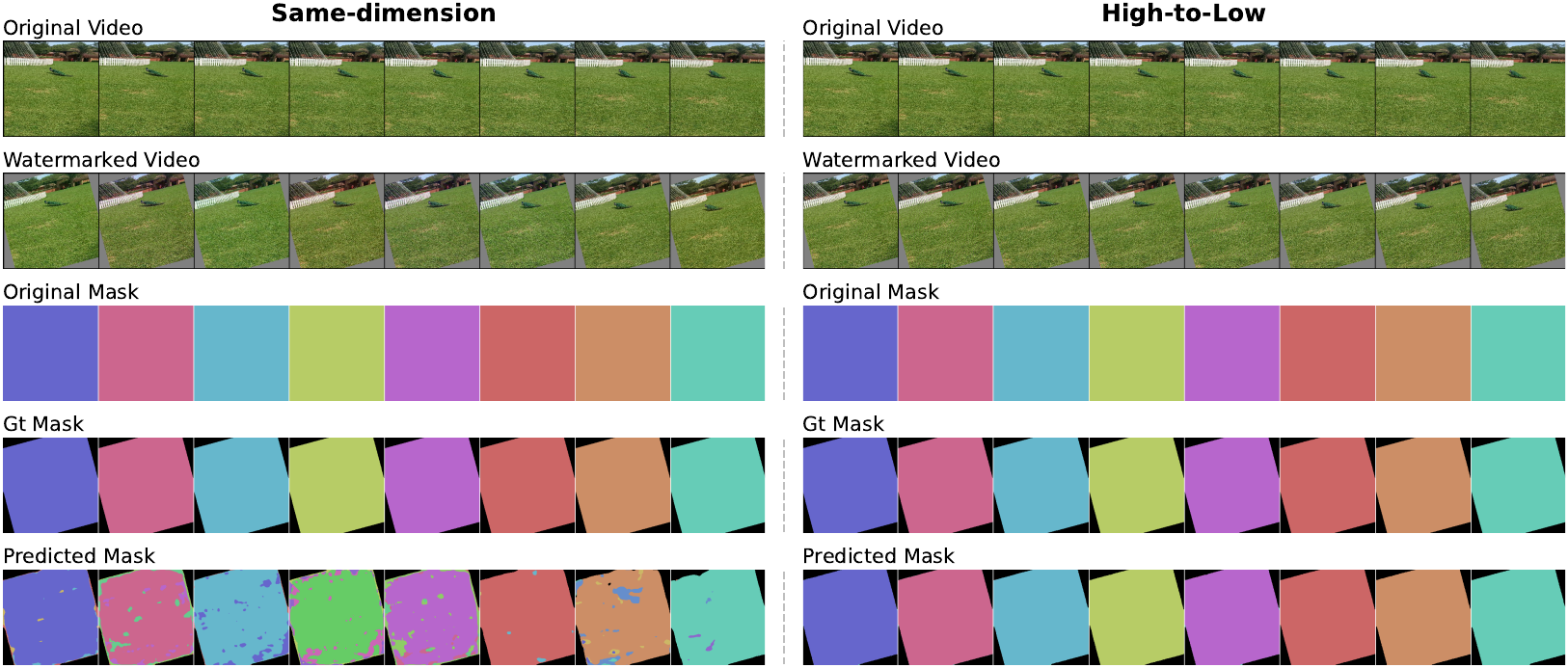}
    \caption{Visualization of predicted multi-channel masks for $\mathcal{M}\{3,3\}$ and $\mathcal{M}\{3,2\}$ under rotation. Different  mask encodings are visualized using distinct colors.}
    \label{fig:multi_rotate}
\end{figure*}

\begin{itemize}
    \item No distortion: see Figure~\ref{fig:multi_clean}.
    \item Under rotation: see Figure~\ref{fig:multi_rotate}.
\end{itemize}


\end{document}